%% file: main.tex
\documentclass[sigconf, authorversion, nonacm]{acmart}

\usepackage{booktabs} % For formal tables
\usepackage{cleveref}
\usepackage{algorithm}
\usepackage{algpseudocode} % For algorithms
\usepackage{xr}

\setcitestyle{numbers}

\input{macros}

\title{
Cross-domain Compositing with Pretrained Diffusion Models}
\author{
Roy Hachnochi$^{1}$ \hspace{0.5cm} 
Mingrui Zhao$^{2}$ \hspace{0.5cm} 
Nadav Orzech$^{1}$ \hspace{0.5cm} 
Rinon Gal$^{1}$ \\
Ali Mahdavi-Amiri$^{2}$ \hspace{0.5cm} 
Daniel Cohen-Or$^{1}$ \hspace{0.5cm} 
Amit Haim Bermano$^{1}$ \and \vspace{2mm} 
$^{1}$Tel Aviv University \hspace{0.5cm} 
$^{2}$Simon Fraser University \and \vspace{2mm} 
\small{\href{https://github.com/cross-domain-compositing/cross-domain-compositing}{https://github.com/cross-domain-compositing/cross-domain-compositing}}
}

\renewcommand{\shortauthors}{Hachnochi, R. et al.}

\begin{document}

\pagenumbering{arabic}
\input{0-abstract.tex}
\input{0-teaser.tex}
\settopmatter{printfolios=true}
\maketitle
\input{1-intro.tex}
\input{2-RW.tex}

\input{3-method.tex}
\input{4-experiments.tex}
\input{5-limitations.tex}
\input{6-conclusions.tex}

\bibliographystyle{ACM-Reference-Format}
\bibliography{bib}

\clearpage
\input{appendix.tex}

\end{document}

%% file: macros.tex
\usepackage{comment}
\usepackage{multirow,bigdelim}
\usepackage{lipsum}
\usepackage{array}
\usepackage{wrapfig}

\usepackage{adjustbox}
\usepackage[percent]{overpic}
\usepackage{makecell}
\usepackage[normalem]{ulem}
\usepackage{blindtext}
\usepackage{xcolor}
\usepackage{soul}

\newcolumntype{C}[1]{>{\centering\let\newline\\\arraybackslash\hspace{0pt}}m{#1}}

\newif\ifdraft
\drafttrue
\ifdraft
\newcommand{\dcc}[1]{{\color{red}[\textbf{DC:} #1]}}
\newcommand{\rgc}[1]{{\color{purple}[\textbf{RG:} #1]}}

\newcommand{\hmc}[1]{{\color{teal}[\textbf{HM} #1]}}

\newcommand{\abc}[1]{{\color{green}[\textbf{AB:} #1]}}

\newcommand{\drop}[1]{}

\else
\newcommand{\dcc}[1]{}
\newcommand{\rgc}[1]{}
\newcommand{\opc}[1]{}
\newcommand{\gcc}[1]{}
\newcommand{\hmc}[1]{}
\newcommand{\abc}[1]{}

\fi

\def\naive{na\"{\i}ve\xspace}

\makeatletter
\DeclareRobustCommand\onedot{\futurelet\@let@token\@onedot}
\def\@onedot{\ifx\@let@token.\else.\null\fi\xspace}

\def\eg{\emph{e.g}\onedot}

\def\ie{\emph{i.e}\onedot}

\makeatother
\usepackage[bottom]{footmisc}
\usepackage{arydshln}

\makeatletter
\def\blfootnote{\xdef\@thefnmark{}\@footnotetext}
\makeatother

\newcommand*{\supref}[1]{appendix \ref{#1}}

%% file: 0-abstract.tex
\begin{abstract}

Diffusion models have enabled high-quality, conditional image editing capabilities. We propose to expand their arsenal, and demonstrate that off-the-shelf diffusion models can be used for a wide range of cross-domain compositing tasks. Among numerous others, these include image blending, object immersion, texture-replacement and even CG2Real translation or stylization. We employ a localized, iterative refinement scheme which infuses the injected objects with contextual information derived from the background scene, and enables control over the degree and types of changes the object may undergo. We conduct a range of qualitative and quantitative comparisons to prior work, and show that our method produces higher quality and realistic results without requiring any annotations or training. Finally, we demonstrate the advantage of cross-domain compositing in data augmentation for downstream tasks.

\end{abstract}

%% file: 0-teaser.tex
%% This is the ``teaser'' command, which puts an figure, centered, below
%%% the title and author information, and above the body of the content.

\begin{teaserfigure}
   \centerline{\includegraphics[width=0.95\textwidth]{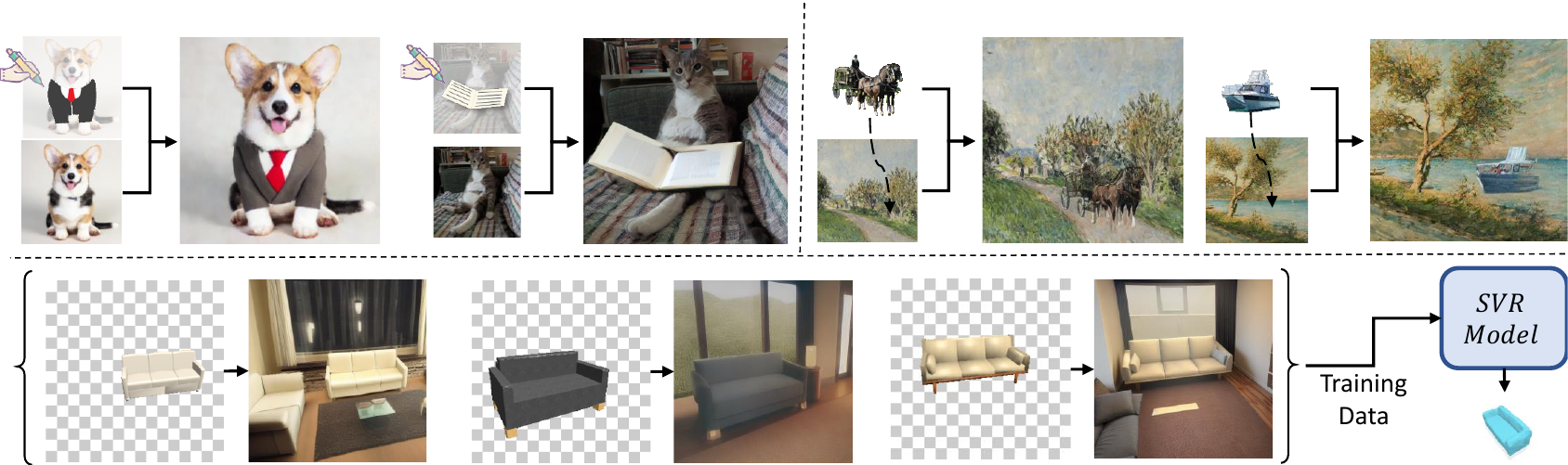}}
  \caption{\textbf{Various applications of our method.} Top-left: local editing of an image by using scribble and text guidance. Top-right: Cross-domain immersion of real image into painting. Bottom: Constructing a geometrically coherent scene around a CG-object and simultaneously altering its style for data augmentation of downstream tasks such as Single View Reconstruction (SVR).}
  \label{fig:teaser}
\end{teaserfigure}

%% file: 1-intro.tex
\section{Introduction}
\label{sec:intro}

The ability to compose new scenes from given visual priors is a long standing task in computer graphics. Commonly, the goal is to extract objects from one or more images, and graft them onto new scenes. This task is already challenging, requiring the novel objects to be harmonized with the scene, with consistent illumination and smooth transitions. However, this challenge is exacerbated when the components originate in different visual domains --- such as when inserting elements from natural photographs into paintings, or when introducing 3D-rendered objects into photo-realistic scenes. In such cases, the objects should be further transformed to match the style of the new domain, without harming their content.

We propose to tackle this Cross-Domain Compositing task by leveraging diffusion-based generative models. Such models have recently lead to remarkable strides in image synthesis and manipulation tasks, where pre-trained diffusion models often serve as powerful visual priors which can ensure realism and coherency while also preserving image content. Here, we aim to use such models to naturally infuse new scenes with cross-domain objects. 

Our approach aims to accomplish two goals: First, it must be capable of inducing \textit{local} changes in an image, based on a visual input. Second, it should allow for minor changes in the input, to better immerse it in the surrounding scene. Ideally, these changes should be \textit{controllable}, allowing a user to dictate the type and extent of similarity they wish to preserve. We tackle these goals by expanding on recent work in diffusion-based inpainting and conditional guidance. We propose an iterative, \textit{inference-time} conditioning scheme, where localized regions of the image are infused with controllable levels of information from given conditioning inputs. 

In \cref{fig:teaser}, we demonstrate a range of applications enabled by this approach, including: injecting real objects into paintings, converting low-quality renders into realistic objects, and scribble-based editing.

We compare our method to existing baselines across multiple tasks and demonstrate its visually pleasing results, to the extent that it can even serve as a data-augmentation pipeline, diminishing the gap between performance on simulated data and real images.

%% file: 2-RW.tex
\section{Related Work}
\label{sec:RW}

\textbf{Image composition} is a well-studied task in computer graphics. Compositions are used for image editing and manipulation \citep{lin2018st}, and even for data augmentation \citep{zhou2022sac,dwibedi2017cut}.
Traditional techniques for inserting 2D object patches into images include alpha matting \citep{smith1996blue} and Poisson image editing \citep{perez2003poisson}. These approaches involve \textit{blending} a foreground object into a background image by creating a transition region. However, compositing typically requires additional transformations, which have recently been tackled in the context of deep learning literature. Some learn to apply appropriate transformations (\eg, scale and translation)~\citep{zhou2022sac,lin2018st} to an object patch to better place it into a scene image. Some works leverage Conditional GANs to map images of different objects (\eg, a basket and a bottle) into a combined sample (\eg, a basket with a bottle in it) that captures the realistic interactions between the objects \citep{azadi2020compositional}. Others improve realism by inserting appropriate shadows~\citep{hong2022shadow}. Another fundamental sub-field is \textit{image harmonization}, which aims to correct for any mismatch in illumination between the new patch and the background~\citep{cong2020dovenet,cun2020improving,tsai2017deep}. \Citep{cong2021bargainnet} do this by encoding illumination information from the background and translating the foreground domain accordingly, while \citep{jiang2021ssh} disentangle image representation into content and appearance, and accomplish harmonization by translating solely the appearance of the foreground by that of the background. Finally, compositing methods that aim at inserting 3D objects into scenes typically first determine the geometry and lighting of the scene. They then generate an image of the 3D object conditioned on these and insert it into the scene~\citep{debevec1998rendering,karsch2011rendering,karsch2014automatic,kholgade20143d,yi2018faces,einabadi2021deep}.

In contrast to these approaches, our work tackles the task of \textit{cross-domain} compositing, where the inserted patch and the scene belong to different visual domains. Our goal is therefore not only object preservation and blending, but also simultaneous matching of domains. Moreover, we leverage pretrained models without having to rely on pretrained detectors or additional labels.

\textbf{Image editing with diffusion models.} The use of diffusion models~\citep{sohl2015deep,ho2020denoising} to modify images is a topic of considerable research. These models have been employed for image inpainting~\citep{sohl2015deep,lugmayr2022repaint,saharia2022palette}, scribble-based modifications~\citep{choi2021ilvr,meng2021sdedit} or in-domain compositing operations~\citep{meng2021sdedit}. More recently, text-based diffusion models~\citep{ramesh2022hierarchical,nichol2021glide,rombach2021highresolutionLDM} have been leveraged for editing operations guided by natural language. These include text or reference-based inpainting~\citep{avrahami2022blendedlatent,nichol2021glide,gal2022image,yang2022paint} and outpainting~\citep{ramesh2022hierarchical}, 
 to directly modifying an image according to given prompts~\citep{kawar2022imagic,valevski2022unitune,mokady2022null,hertz2022prompt,tumanyan2022plug}.

Our method is related to the guided inpainting line of work, where the inpainting is guided using a visual reference. However, in contrast to current approaches, the visual reference need not be from the same visual domain. Moreover, our work employs pretrained diffusion models, without requiring task-specific training~\citep{yang2022paint,saharia2022palette,nichol2021glide} or tuning on individual reference images~\cite{gal2022image}.

\textbf{Single-view 3D reconstruction (SVR)} is the task of reconstructing a 3D model from a single 2D image. One major challenge in this domain is that existing deep learning solutions, which heavily rely on synthetic data \cite{ShapeNet} for training, struggle when applied to real-world images. Existing works \cite{choy20163d,wu2017marrnet,wu2022magicpony, collins2022abo} attempted on this issue by fine-tuning SVR networks over in-the-wild images, pre-segmenting objects from the backgrounds or augmenting training data background by naive composition, while none of them explicitly tackle the sim-to-real domain gap between synthetic training dataset and real-world testing images. In this paper, we demonstrate that our composition method can serve as an effective augmentation technique to address this domain gap by seamlessly generating a realistic and plausible scene around a rendered 3D model.

%% file: 3-method.tex
\begin{figure*}[t!]
\centerline{\includegraphics[width=0.9\textwidth]{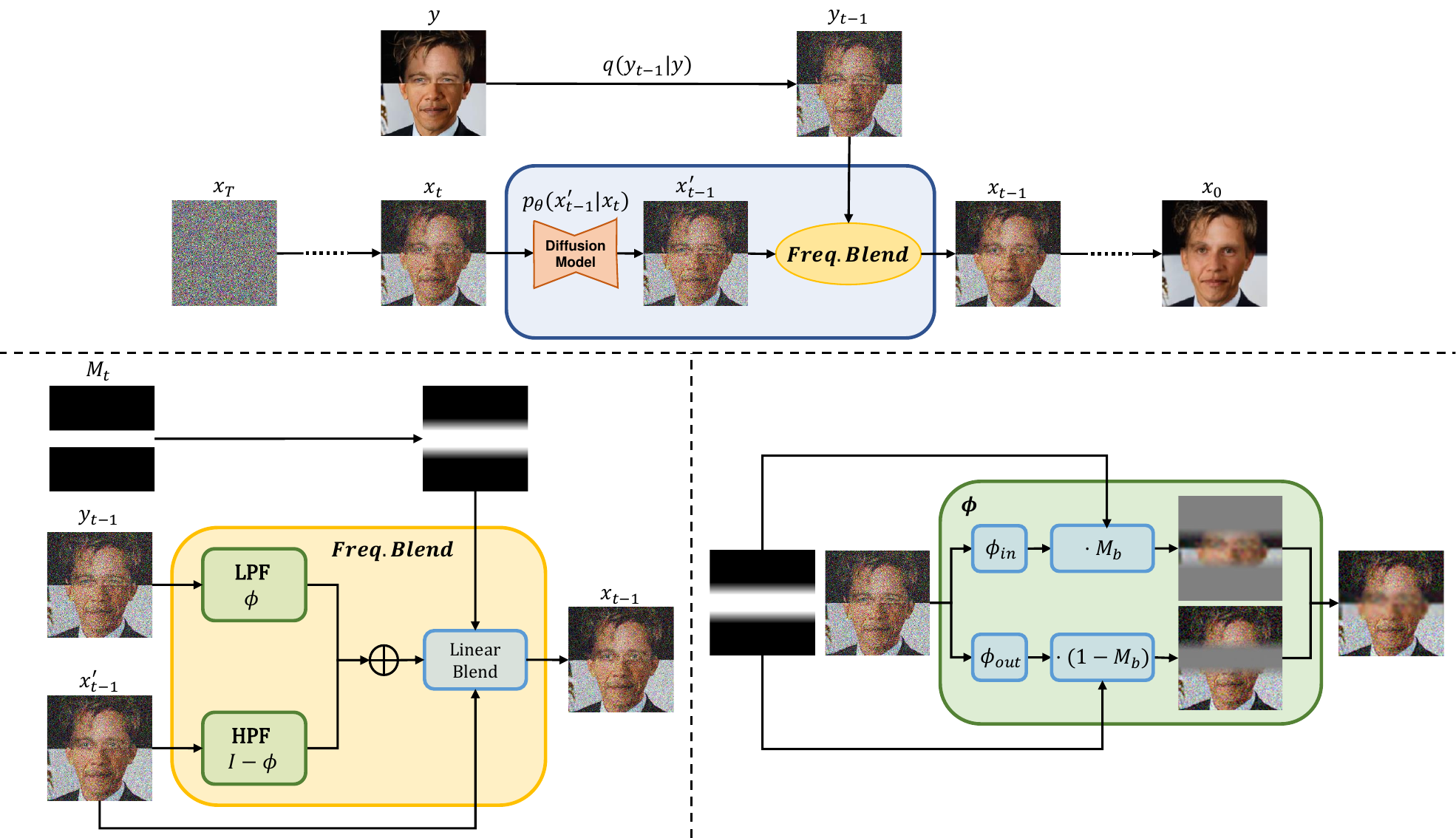}}
  \caption{\textbf{Single diffusion step of our method.} Top: In each step the image is denoised using the diffusion model, and blended with the noised reference image using our method. Bottom left: The image is refined by mixing low frequencies of the reference image with high frequencies of the current step image, using the same LPF. The result is blended with the current image according to the time mask. Bottom right: Our blending algorithm allows control over the filters in different image regions and blends the result by using smoothed mask.}
  \label{fig:method}
\end{figure*}

\section{Method}
\label{sec:method}

\subsection{Preliminaries}\label{ss:preliminaries}
Diffusion models~\citep{ho2020denoising} have advanced the state-of-the-art in image generation \cite{dhariwal2021diffusion}. 

They work by learning to reverse a time-dependent noising process, given by the closed form equation:
\begin{align}\label{eq:DDPM_q}
\begin{split}
&q_\theta(x_t|x_{t-1}) = \mathcal{N}(\sqrt{1-\sigma_t}x_{t-1},\sigma_t\textbf{I}), \\
&x_t=\sqrt{\overline{\alpha}_t}x_0+\sqrt{1-\overline{\alpha}}_t\epsilon
\end{split}
\end{align}
where $x_0$ is a given, noise-less image, $t$ is a timestep, $\epsilon \sim \mathcal{N}(0,\textbf{I})$ is a random noise sample, $\sigma_t$ is a predefined noise-level schedule,  $\overline{\alpha}_t=\Pi_{s=1}^t\alpha_s$ and $\alpha_t=1-\sigma_t$.

At inference time, diffusion models can be used to synthesize new images by taking a random noise sample $x_T \sim \mathcal{N}(0,\textbf{I})$ and iteratively denoising it. Given a noised image $x_t$ at time-step $t$, the model predicts the next-step $x_{t-1}$ as: 
\begin{align}\label{eq:DDPM_p}
\begin{split}
&p_\theta(x_{t-1}|x_t) = \mathcal{N}(\mu_\theta(x_t,t),\sigma_t\textbf{I}), \\
&\mu_\theta(x_t,t)=\frac{1}{\sqrt{\alpha_t}} \left( x_t-\frac{1-\alpha_t}{\sqrt{1-\overline{\alpha}_t}}\epsilon_\theta(x_t,t) \right),
\end{split}
\end{align}
where $\epsilon_\theta$ is a neural network that learns to predict the noise at each step. Additionally, at each timestep, one can approximate the noiseless image $\hat{x}_0$ through:
\begin{align}\label{eq:DDPM_x0}
\begin{split}
\hat{x}_0=\frac{x_t}{\sqrt{\overline{\alpha}_t}}-\frac{\sqrt{1-\overline{\alpha}_t}\epsilon_\theta(x_t,t)}{\sqrt{\overline{\alpha}_t}}.
\end{split}
\end{align}

This method may be further generalized for sampling from conditional distributions by conditioning the denoising network $\epsilon_\theta(x_t,t,y)$ and applying classifier-free guidance \cite{ho2022classifier}. $y$ may be any type of condition, such as a class label or a text-embedding derived from an LLM such as CLIP \cite{radford2021learning}.

\subsection{Masked ILVR}\label{ss:masked_ILVR}
Our goal is to create composite images which contain parts from different visual domains. To do so, we require a method that allows for retaining the structure and semantics of the object, while changing its appearance to match the new domain. The result should ideally both appear similar to the reference image, and realistically match the background domain. Moreover, this method should tackle other compositing requirements, such as realistic blending.

We propose to tackle all these requirements at once at \textit{inference-time} by leveraging a pretrained diffusion model. To do so, we would like to condition the diffusion process on external image inputs, such as a desired background and a set of foreground objects. One way to do so is to leverage ILVR~\cite{choi2021ilvr}: for each denoising step $t$, the diffusion model first predicts a proposal $x'_{t-1} \sim p_{\theta}(x'_{t-1}|x_t)$. Then, the proposal is refined by injecting low-frequency data from a reference image via the following update rule:
\begin{align}\label{eq:ILVR}
\begin{split}
x_{t-1} &= \phi(y_{t-1})+(I-\phi)(x'_{t-1}) ,\\
&= x'_{t-1}+\phi(y_{t-1})-\phi(x'_{t-1}) ,
\end{split}
\end{align}
where $y_{t-1} \sim q(y_{t-1}|y_0)$ is a noised version of the conditioning image (sampled through \cref{eq:DDPM_q}), and $\phi$ is a low-pass filter operation. Intuitively, this process overwrites all the low-frequency information with that derived from the conditioning image, allowing for a degree of structure to be inherited while still enabling high-frequency modifications. In practice, $\phi$ is implemented through a bi-linear down-sampling and up-sampling step with a scaling factor of $N$. This refinement operation is repeated for every diffusion step until some user-defined threshold $T_{stop}$, after which denoising continues without guidance for every $t<T_{stop}$. These two parameters, $N$ and $T_{stop}$, grant control over similarity to the reference image. Higher $N$ and higher $T_{stop}$ will result in higher diversity but less fidelity to the conditioning image, since fewer frequencies are overridden and for fewer steps. These parameters therefore introduce control over the inherent trade-off between \textit{fidelity} to the reference image and its \textit{realism}. We aim to expand this idea to provide the user with \textit{local} control over this trade-off.

\begin{figure}[t!]
\setlength{\belowcaptionskip}{-8pt}
\centerline{\includegraphics[width=0.8\linewidth]{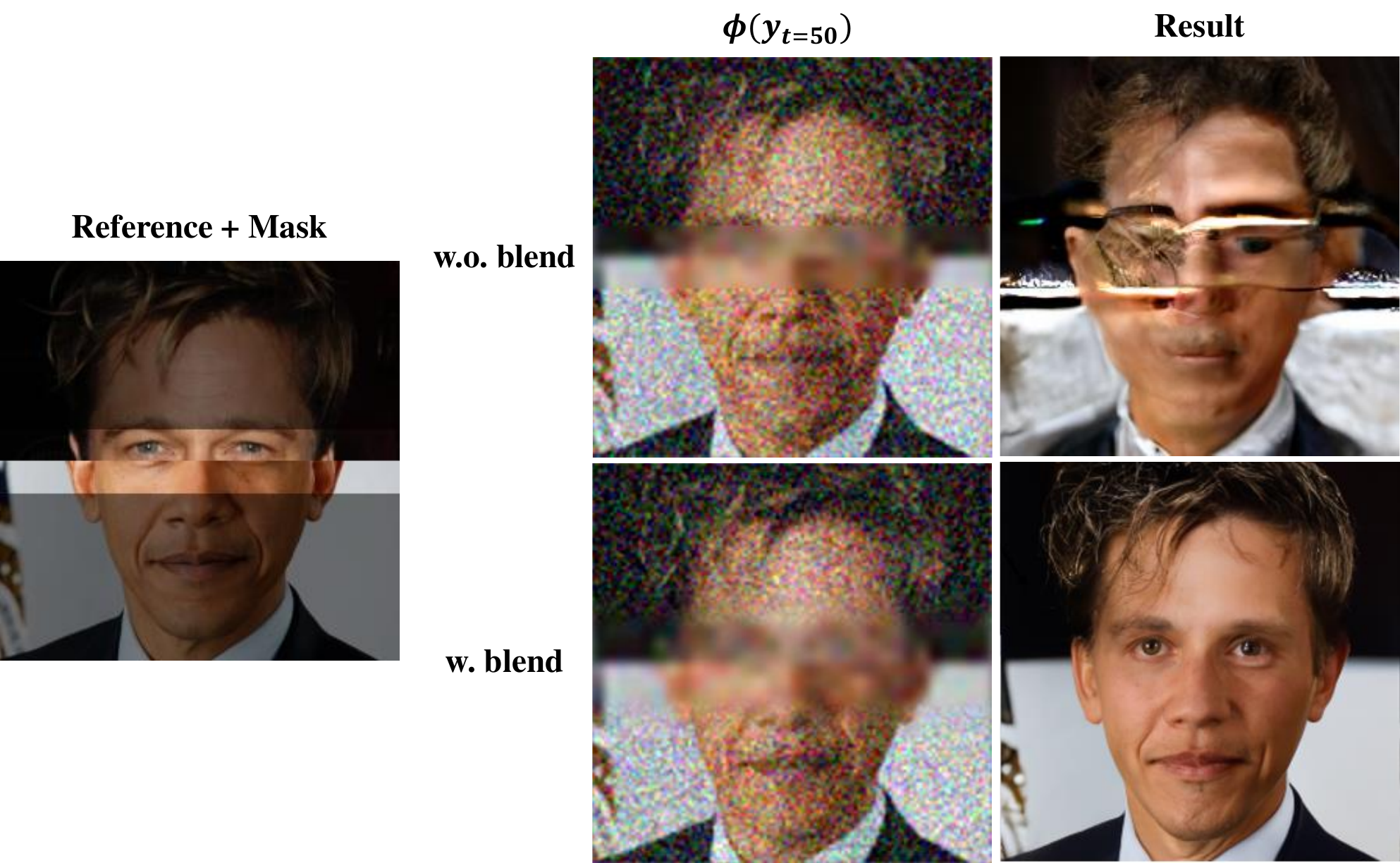}}
  \caption{Demonstration of the aliasing artifacts and an intermediate timestep of $\phi(y_t)$ for $t=50$. We hypothesize that the sharp edges introduced at intermediate steps are the cause of these artifacts. Our blending method mitigates these artifacts.}
  \label{fig:aliasing}
\end{figure}

A \naive way to leverage ILVR for localized editing is to apply it inside local masks. There, we can control the ILVR parameters $T_{stop}$ and $N$ individually. However, ILVR was designed as a global method, and attempts to apply it in a local manner lead to the advent of visual artifacts (see \cref{fig:aliasing}). We investigate these artifacts and observe that they appear along the boundaries of the masked regions, and become more pronounced as the gap increases between the noise statistics of the image regions (\eg, when $N$ changes significantly between regions). We hypothesize that this is a consequence of frequency aliasing~\cite{Karras2021alias,zhang2019shiftinvar}, originating from the sharp transition between regions. To overcome these hurdles, we propose to modify ILVR with the following scheme, illustrated in \cref{fig:method}.

\textbf{Localizing the control.} To allow local control, we apply the ILVR process into separate image regions. Given a reference image $y \in \mathbb{R}^{H \times W \times C}$ and a mask $M \in [0, 1]^{H \times W}$, the user may separately define $N_{in}, N_{out}$ for the low-pass filters $\phi_{in}, \phi_{out}$ respectively, and $T_{in}, T_{out}$ as the ILVR strengths per region. These allow us to define a combined linear low-pass filtering operator:
\begin{align}\label{eq:blend}
\phi(x;M_b)=M_b\phi_{in}(x) + (1 - M_b)\phi_{out}(x),
\end{align}
where $M_b$ is the blending mask. Following ILVR, we would also like to enable different conditioning-stop times in each region. This is done by introducing a `time-mask' $M_t$ into \cref{eq:ILVR}:
\begin{align}\label{eq:maskedILVR}
x_{t-1} = x'_{t-1}+M_t(\phi(y_{t-1})-\phi(x'_{t-1})) ,
\end{align}
where $M_t$ is set to $0$ in regions for which $t < T_{stop}$.
Formally, the blending mask $M_b$ and the time mask $M_t$ are defined as:
\begin{align}\label{eq:masks}
\begin{split}
&M_b=M ,\\
&M_T=(1-T_{in})T \cdot M+(1-T_{out})T \cdot (1-M) ,\\
&M_t^{(i,j)}(t)=\begin{cases}
     0 & :t<M_T^{(i,j)} \\
     1 & :t \geq M_T^{(i,j)}
     \end{cases} ,
\end{split}
\end{align}
where $T$ is the total number of diffusion steps and $T_{in}$, $T_{out}$ define the fraction of steps that should use the ILVR condition in each region. For example, $T_{in}=0.2, T_{out}=1$ would result in applying ILVR guidance for 20\% of the diffusion steps inside the mask and 100\% outside. This per-region parameter separation allows for local control over the fidelity-realism trade-off, independently within each area of the image. We note that this control may also be applied to latent diffusion models \cite{rombach2022high} if the latent space has spatial consistency with the pixel space. In such cases, we can simply apply low-pass filtering in latent space.

\textbf{Overcoming the aliasing artifacts.} As noted, we find that for pixel space diffusion models (\eg, Guided Diffusion \cite{dhariwal2021diffusion}) this method induces unnatural, wave-like artifacts on the resulting image. To mitigate them, we propose two alternatives that aim to remove the sharp boundaries in the noise statistics.

As a first alternative, we propose to apply the ILVR blending steps (\ie, \cref{eq:blend,eq:maskedILVR}) in the approximated $\hat{x}_0$-space instead of $x_t$. When doing so, the low-pass filter $\phi$ is applied only to the image and not to the noise, which can then be added to the image post-blending. This ensures that no sharp edges are existent in the noise map, maintains the appropriate noise levels for each timestep, and successfully overcomes the artifacts.

A second alternative is to add a smoothing operator $b(M)$ to be applied on the masks $M_b, M_t$. This approach can overcome the aliasing effect by smoothing transitions between different regions of the images. We observe experimentally (\supref{ss:aliasing}) that the blurring factor should have a spatial length of roughly $max(N_{in}, N_{out})$. That is, it should ensure that the transition remains smooth even in the down-scaled images. This eliminates the sharp transitions of the noise levels between image regions, and eliminates the noise, as shown in \cref{fig:aliasing}. Additional analysis and implementation details of this approach are given in \supref{ss:smoothing} and \supref{ss:aliasing}.

In practice, we find that the first method works better when using a model which is able to predict a plausible $\hat{x}_0$ image in intermediate diffusion steps, such as models trained on more constrained domains. We hypothesize that this is related to the ability to meaningfully interpolate between the reference image and the prediction. If the prediction is out-of-domain, the interpolation will deteriorate the image, and result in non-meaningful guidance. In such cases, it is preferable to use the smoothed noise-space blending approach.

Finally, we note that for latent diffusion models this effect is more subdued. We theorize that this is due to the latent decoder's abilities to compensate for some of the artifacts in the diffused latents.

\textbf{Step repetitions.} Our approach allows for cross-domain information to diffuse between the two image regions, allowing for a degree of domain-matching. However, the rate of information transfer is largely limited by the receptive field of the denoising network which in some cases may be too low (\eg, for compositing large objects). We would therefore like to infuse the object with more cross-domain information from the background, without changing it or causing it to lose structure. RePaint~\cite{lugmayr2022repaint} showed that simply repeating denoising steps can have such an effect, in essence extending the receptive field and allowing more time for information to pass between the regions. We thus employ a similar approach. We note that dedicated inpainting models can also provide an advantage for structure preservation when such are available. However, these are not designed to perform inpainting with a given visual object, while our method does.

%% file: 4-experiments.tex
\section{Applications and Evaluation}
\label{sec:experiments}

We demonstrate various applications of our method and perform qualitative and quantitative evaluations and comparisons. In \cref{ss:modification} we present results on image modification guided by a rough user edit, including comparisons with prior baselines. \Cref{ss:immersion} shows how our compositing method may be applied to object immersion, and how it compares with state-of-the-art approaches. In \cref{ss:SVR} we utilize our method to bridge the domain gap between rendered and photo-realistic images which are then used to train Single View 3D-Reconstruction (SVR) models. We show that training on such data can lead to improved downstream performance on real images. In \cref{ss:ablation} we conduct an ablation study and demonstrate the effects of the model's parameters. Throughout these experiments we use Stable Diffusion \cite{rombach2021highresolution} as our diffusion model. For additional results see \supref{ss:additional}.

\subsection{Evaluation metrics.} \label{ss:metrics}
We evaluate our method on two aspects: fidelity to the original images, and realism. For fidelity analysis, we use LPIPS \cite{zhang2018perceptual} and PSNR to measure the similarity between the foreground object pre- and post-composition. For realism, we use two metrics: First, the CLIP directional similarity score (\cref{eq:clip_dir_sim}), introduced in \cite{gal2021stylegannada} and established as a metric in \cite{brooks2022instructpix2pix}. Here, we compare the pre- and post-composition CLIP-space embeddings of the foreground object and ensure they match a given textual direction (\eg, "photo" to "painting"). Second, we use a CLIP similarity improvement metric (\cref{eq:clip_sim_imp}), where we investigate if the CLIP-similarity between the object and the background domain has improved post-composition. These metrics are defined by:
\begin{align}\label{eq:clip_dir_sim}
\begin{split}
\begin{gathered}
\Delta T = E_T(t_{target}) - E_T(t_{source}), \; \Delta I = E_I(I_{pred}) - E_I(I_{ref}), \\
CLIP_{dir} = \frac{\Delta I \cdot \Delta T}{|\Delta I||\Delta T|} ,
\end{gathered}
\end{split}
\end{align}
\begin{align}\label{eq:clip_sim_imp}
CLIP_{SI} = \frac{E_I(I_{pred}) \cdot E_T(t_{target})}{E_I(I_{ref}) \cdot E_I(I_{target})} ,
\end{align}
 where $E_T, E_I$ are CLIP's textual and image encoders, respectively, and $t_{target}, t_{source}$ are the texts describing the target and source domains. Last, we evaluate background preservation using LPIPS and PSNR. 

\subsection{Image modification via scribbles} \label{ss:modification}
We investigate the use of our method for scribble-based editing, where only a user-designated area should change based on the scribble and some text prompt.

\begin{figure}[b!]
\setlength{\belowcaptionskip}{-10pt}
   \centerline{\includegraphics[width=0.97\linewidth]{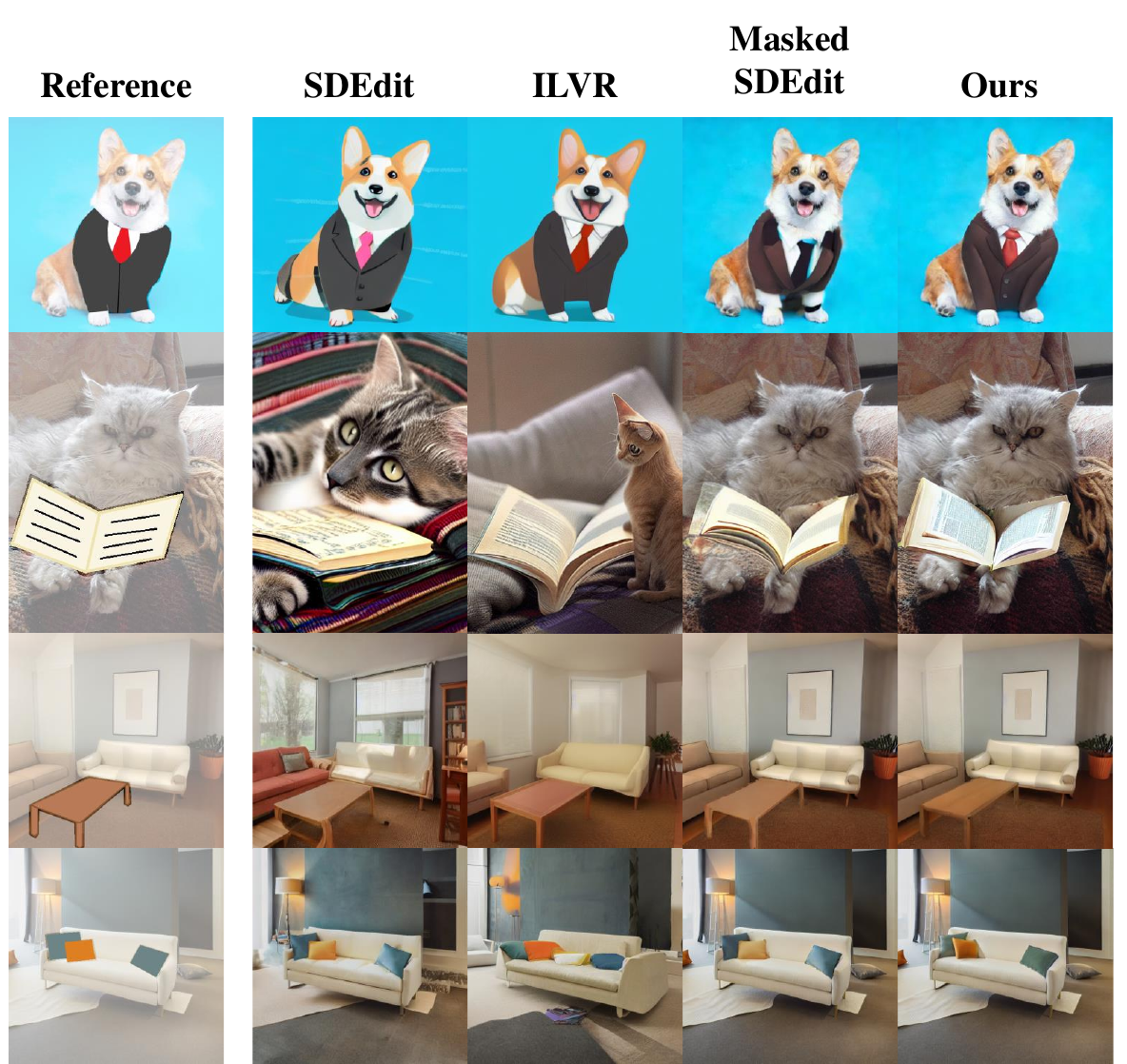}}
  \caption{\textbf{Qualitative comparison of local image editing using scribble guidance.} For each method we run a parameter sweep and choose the empirically best results. Other methods fail in preserving the background or in altering the scribble into a realistic object. Specifically notice the high-frequency details which were added to the scribbles, such as folds and buttons to the suit or shadows and texture to the pillows. Textual prompts from top to bottom: \textit{a corgi wearing a business suit}, \textit{a cat reading a book}, \textit{a living room with a sofa and a table}, \textit{a sofa with pillows}.}
  \label{fig:exp-scribbles}
\end{figure}

\textbf{Qualitative comparison.} In \cref{fig:exp-scribbles} we compare our method qualitatively to SDEdit and ILVR, including a masked-SDEdit approach. SDEdit and ILVR struggle at preserving the image background. For high editing strength (high $strength$ for SDEdit, high $T_{stop}$ for ILVR) their results match the text prompt but have no similarity to the reference image, while low edit strengths result in failure to transform the scribbles to realistic objects. This is natural since these are both global guidance methods. Masked-SDEdit succeeds in preserving the image but often fails in morphing the scribbles. Our method succeeds both in preserving background features and in adherence to the scribbled guidance. Moreover, our results are altered to better fit the background. These are a consequence of the increased control over the amount of overridden frequencies and separated foreground-background control.

\textbf{Quantitative comparison.} We evaluate all methods over a set of $24$ scribble images, using the metrics of \cref{ss:metrics}. Results are shown in \cref{tab:metrics} (top). For CLIP-based scores we use $t_{source}="drawing", \; t_{target}="photo"$. Our method better matches the target domain and improves background preservation, but has lower fidelity scores. This is due to the comprehensive domain changes, which modify the content in a way that harms pixel-level scores. 

\subsection{Object Immersion} \label{ss:immersion}

\begin{figure}[h!]
   \centerline{\includegraphics[width=1.0\linewidth]{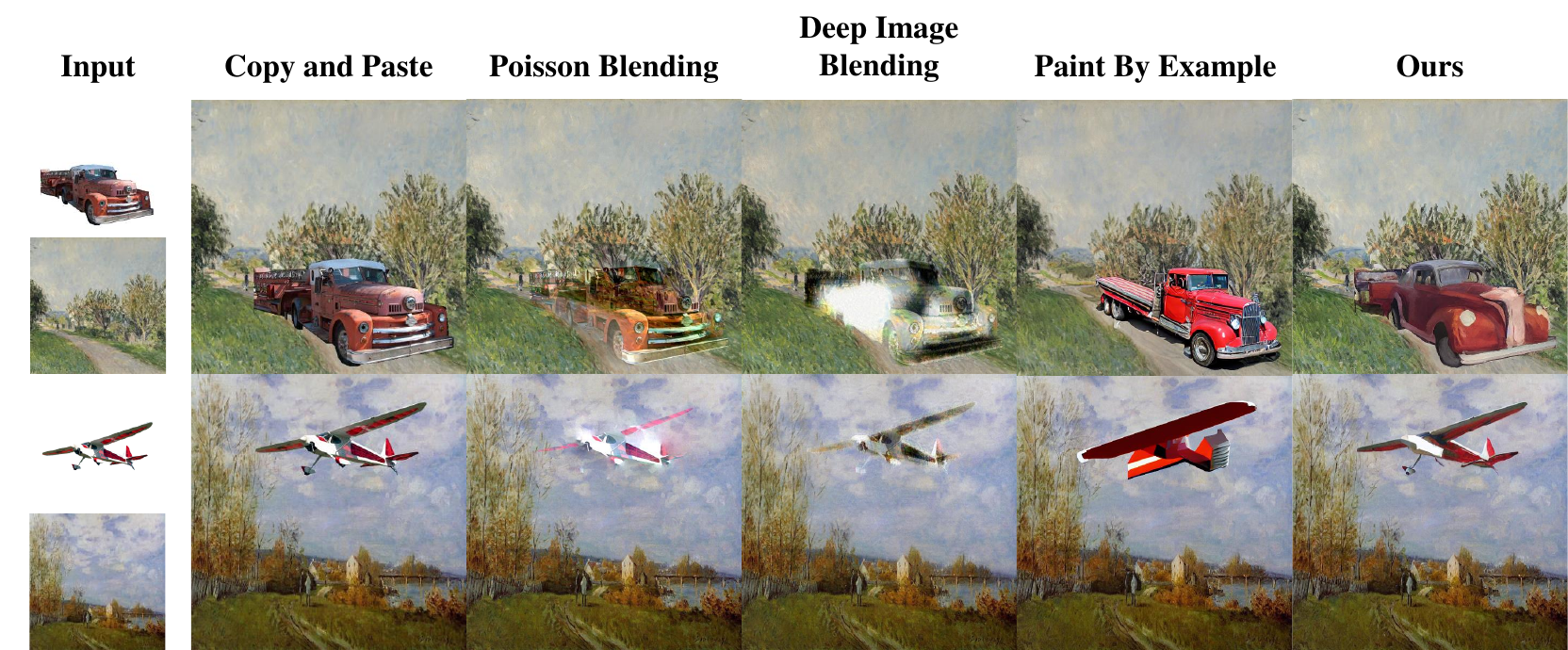}}
  \caption{\textbf{Object immersion comparison.} We compare our method on the task of object immersion across several methods. Our results (right column) exhibit more adequate style-matching, crafting the object to match its background while leaving it intact.}
  \label{fig:exp-immersing}
\end{figure}

The goal of object immersion is to blend an object into a given styled target image, (\eg a painting) while preserving the coarse features of the object and transforming only its style to that of the target. 
%We utilize our cross-domain compositing method for this task by first pasting the cropped source object onto the target image, and applying our method to infuse its coarse details while allowing finer details to be modified.

\begin{table*}[tb!]
\caption{\textbf{Quantitative comparison.} We evaluate fidelity and realism for the foreground, as well as background preservation. Our method demonstrates the best balance between fidelity and realism, and achieves the best background preservation scores among the other latent-diffusion-based methods.}
\begin{tabular}{@{}cccccccc@{}}
\toprule
& & PSNR FG $(\uparrow)$ & LPIPS FG $(\downarrow)$ & PSNR BG $(\uparrow)$ & LPIPS BG $(\downarrow)$ & $CLIP_{dir}(\uparrow)$ & $CLIP_{SI} (\uparrow)$ \\ \midrule
\multirow{4}{*}{Scribbles} & SDEdit & 25.596 & 0.399 & 17.880 & 0.431 & 0.054 & 1.046 \\
& ILVR & \textbf{26.015} & \textbf{0.364} & 19.047 & 0.383 & 0.058 & 1.059 \\
& Masked SDEdit & 25.562 & 0.379 & 25.504 & 0.215 & 0.076 & 1.061 \\
& \textbf{Ours} & 25.059 & 0.394 & \textbf{29.526} & \textbf{0.068} & \textbf{0.093} & \textbf{1.074} \\ \midrule
\multirow{4}{*}{Immersion} & Poisson Image Blending & 27.387 & 0.439 & \textbf{40.233} & \textbf{0.037} & 0.019 & 1.135 \\
& Deep Image Blending & 27.742 & 0.345 & 38.297 & 0.046 & -0.003 & 1.110 \\
& Paint By Example & 24.128 & 0.535 & 20.789 & 0.208 & -0.030 & 1.027 \\
& \textbf{Ours} & \textbf{29.683} & \textbf{0.339} & 23.064 & 0.159 & \textbf{0.028} & \textbf{1.214} \\ \bottomrule
\end{tabular}
\label{tab:metrics}
\end{table*}

\begin{figure}[]
   \centerline{\includegraphics[width=1.0\linewidth]{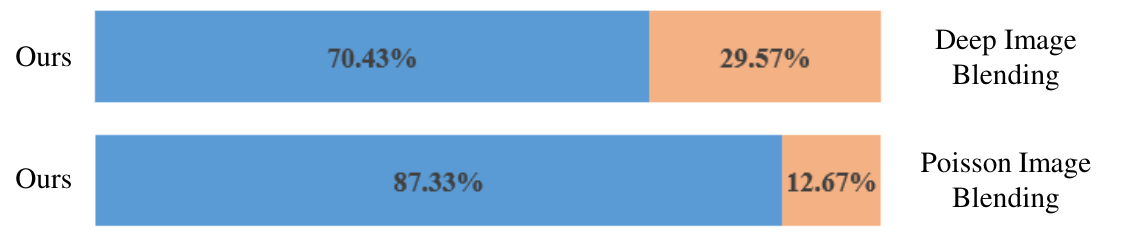}}
  \caption{\textbf{Object immersion user study.} Participants were presented with input images and pairs of outputs from our method and a baseline and asked to select the result which best matches the object to the background, while preserving its characteristics. We show the percentage of users who preferred each method.}
  \label{fig:immersion_us}
\end{figure}

\textbf{Quantitative evaluation.} We compare our method with Deep Image Blending \cite{zhang2020deep}, Poisson Image Blending \cite{perez2003poisson}, and Paint By Example \cite{yang2022paint}, a concurrent work that trains a Diffusion Model to allow subject-driven image inpainting, based on a single reference image. Qualitative results are provided in \cref{fig:exp-immersing}. For our method we choose $T_{in}=0.5, N_{in}=2, R=0.2$ (see \cref{ss:ablation}), providing a reasonable trade-off between fidelity to the reference image and style-matching, and $T_{out}=1,N_{out}=1$ to perfectly preserve the background. We manually curate a 40-image evaluation set, constructed by pasting cropped objects of various sizes from \citep{lin2014microsoft} on top of paintings from \citep{Landscapes-kaggle}. We evaluate our method through two approaches. First, we conducted a user study where users were asked to pick the best results between pairs of our result and one of the compared methods. We collected answers from $116$ participants, each answering $20$ questions, for a total of $2,320$ comparisons. The results are provided in \cref{fig:immersion_us}. Our method outperforms the baselines on user preference scores. In \cref{tab:metrics} we show the fidelity and realism metrics. For the CLIP-based scores, we use $t_{source}="photo", \; t_{target}="oil \; painting"$. Our method outperforms all baselines in foreground fidelity and realism scores, but shows diminished background preservation. Further examination shows that this is the result of the use of a latent diffusion model, which contains an encoder-decoder setup. This approach compresses the image and therefore loses some information. To better quantify our background preservation, we thus also evaluate a reconstruction-only setup. We report reconstruction background PSNR and LPIPS results as $23.478$ and $0.181$ respectively, which align with our scores (23.064, 0.159, \cref{tab:metrics}), and with Paint By Example (20.789, 0.208), which is also an LDM-based method.

\subsection{Single-view 3D Reconstruction (SVR)} \label{ss:SVR}
Current SVR works struggle when applied to real-world images, due to training on synthetic data but inferring on real images. Our method stands as a powerful tool for bridging this domain gap. We demonstrate this on rendered shapes from ShapeNet \cite{DISN} by generating backgrounds with Stable Diffusion's inpainting model \cite{rombach2021highresolutionLDM}. Using our method, we maintain control over the degree of domain matching between the foreground and augmented background using $T_{in}$. We focus our experiments on two model types: sofas and chairs. For each, we generate augmented datasets for different $T_{in}$ values. In addition, we prepare baseline data by \naive image composition (copy \& paste) with background images generated using Stable Diffusion. Subsequently, we trained an off-the-shelf SVR network, $\text{D}^2\text{IMNet}$ \cite{li2021d2im}, on each of these datasets, and assessed it against real in-the-wild images. Note that $T_{in} = 1$ in our method reduces to vanilla inpainting and prevents changes to the foreground. SVR models trained on $T_{in} = 1$, original and copy \& paste datasets together serve as baselines.

\begin{figure}[t!]
   \centerline{\includegraphics[width=0.96\linewidth]{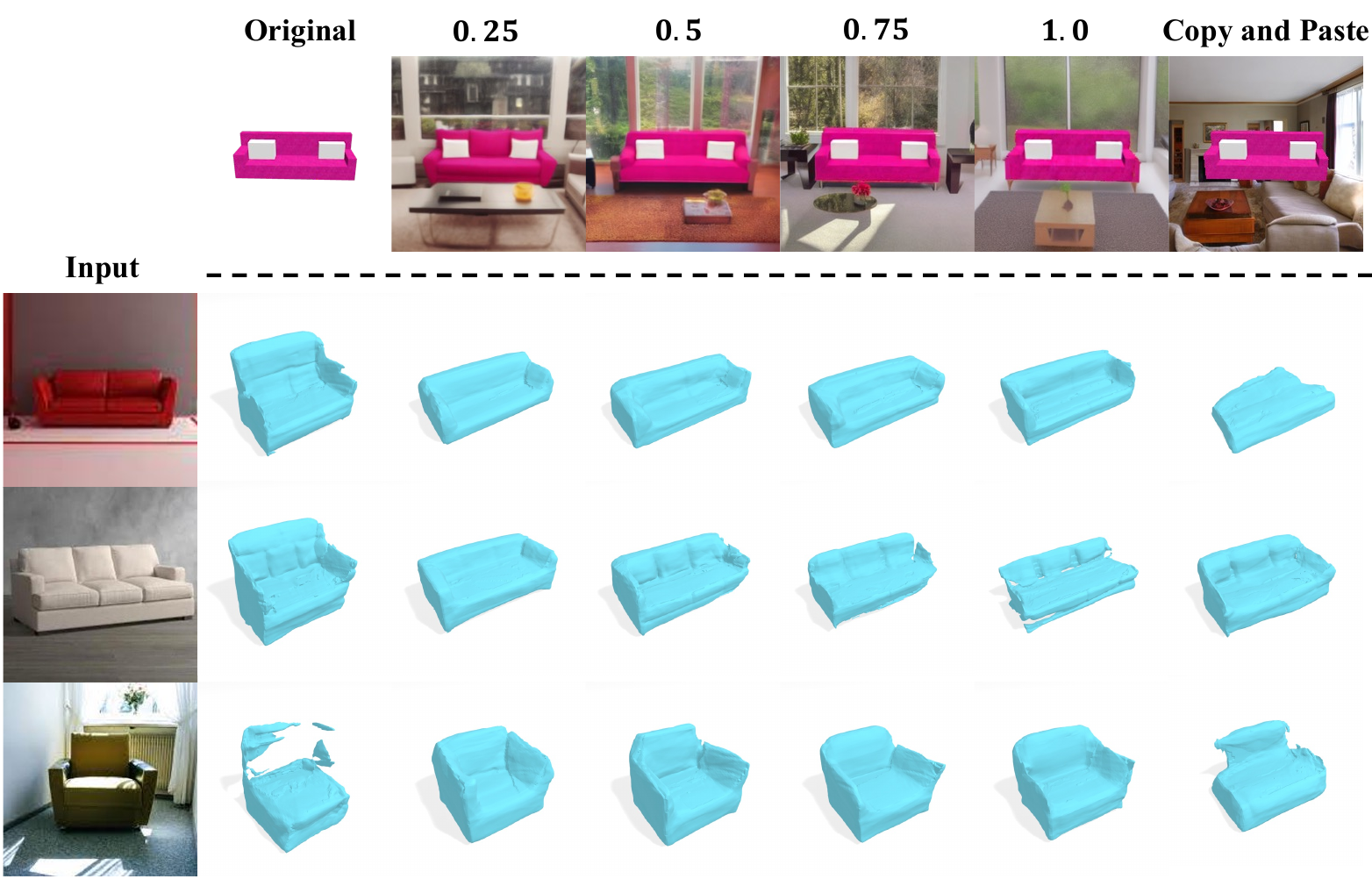}}
  \caption{\textbf{Single View 3D Reconstruction results of $\text{D}^2\text{IMNet}$ trained on different dataset.} Top: Sample image from each dataset. Bottom: SVR results on in-the-wild images produced by respective SVR model. Using our method as augmentation proves to generalize better to real images.}
  \label{fig:SVR-samples}
\end{figure}

\textbf{Quantitative evaluation.} We collect 100 sofa and chair images from the internet with typical poses, no obstructions, and varying backgrounds and lighting conditions. \Cref{fig:SVR-samples} shows qualitative SVR results across different augmentation setups. Both \naive baselines (original, and copy \& paste) fail on real images. We quantify our advantage through a 2D IoU \cite{zou2023segment} metric between the input object mask and the reconstructed 3D entity silhouette rendered at the corresponding camera pose, shown in \cref{tab:2DIoU}, for more details see \supref{ss:2DIoU}. Our augmentation method improves the baselines by: (a) generating a realistic and contextual-aware background, (b) transforming the objects from low-quality renders to more photo-realistic results. Please refer to \supref{ss:additional} for additional results and discussion.

% \begin{table}[]
% \caption{Chamfer $L_1$ distance evaluation on SVR models tested on different dataset. Original model and 0.5 model stands for SVR model trained on the original dataset and $T_{in} = 0.5$ dataset, respectively.}
% {\footnotesize
% \begin{tabular}{@{}ccc@{}}
% \toprule
%          & Original Model & 0.5 Model    \\ \midrule
% Original Data & 0.0743   & 0.0826 \\ \midrule
% 0.5 Data   & 0.167    & 0.0817 \\ \bottomrule
% \end{tabular}
% }
% \label{tab:cross_test}
% \end{table}

\begin{table}[]
\caption{2D IoU evaluated on models trained on different datasets. The numbers in the first row describe the $T_{in}$ value used for background augmentation. CP stands for copy-paste dataset.}
\begin{tabular}{@{}ccccccc@{}}
\toprule
 & Original & 0.25 & 0.5 & 0.75 & 1.0 & CP \\ \midrule
Sofa           & 0.535   & \textbf{0.650}        & 0.623       & 0.554        & 0.464       & 0.486     \\ \midrule
Chair          & 0.434 & 0.529   &\textbf{ 0.564} & 0.474  & 0.455  & 0.301 \\ \bottomrule

\end{tabular}
\label{tab:2DIoU}
\end{table}

\begin{figure}[t!]
   \centerline{\includegraphics[width=1.0\linewidth]
   {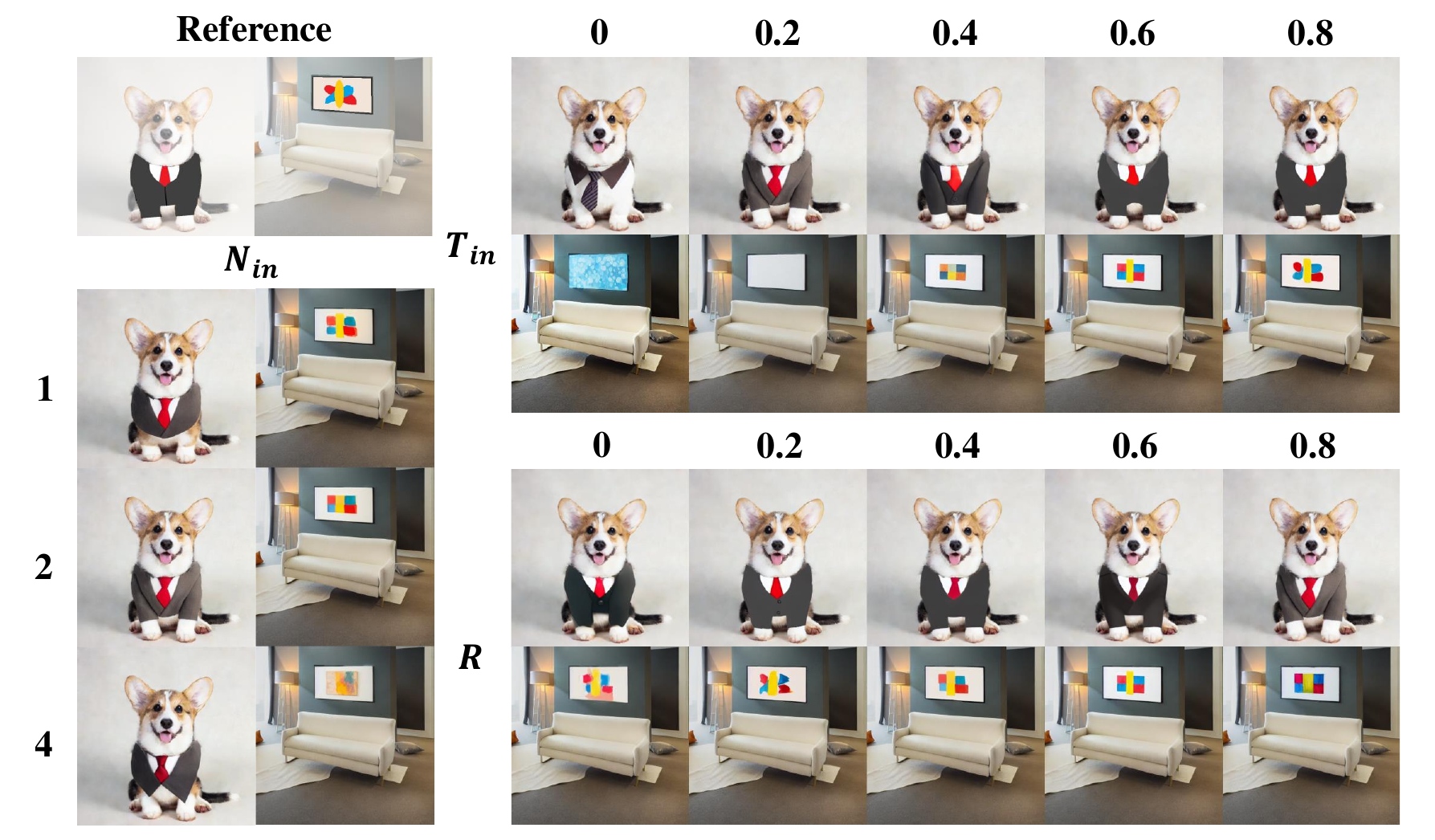}}
  \caption{\textbf{Effect of parameters.} $N_{in}$ is the foreground scaling factor, and controls the structural similarity to the reference object. $T_{in}$ is the strength of ILVR for the foreground, and controls fine-grained changes of the object. $R$ is the relative timestep from which re-sampling begins, allowing details and style from the background to propagate into the object.}
  \label{fig:exp-ablation}
\end{figure}

\subsection{Parameter Ablation} \label{ss:ablation}
In \cref{fig:exp-ablation} we inspect the effect of each parameter on the results. We denote $T_{in}$ as the portion of diffusion steps for which our method is applied in the masked region, $N_{in}$ as the scaling factor for the masked region and $R$ as the relative diffusion step from which re-sampling is performed (see \supref{ss:repaint}). For each parameter we fix the other two and investigate its effect. Lower $T_{in}$ results in less similarity to the reference image, allowing more details to be added, with the extreme case of $T_{in}=0$ reducing to simple inpainting. This results in less-appealing and less-accurate outputs, showing that naive inpainting with textual guidance is not sufficient for these edits. Increasing $N_{in}$ has a similar effect, while using a large $N_{in}$ may result in breaking the structure of the object. $R$ controls the effective receptive field of the guided inpainting, such that increasing it allows for more semantic and style details to be transferred into the edited region. In \supref{ss:metrics-analysis} we provide a deeper analysis of these effects, including a quantitative evaluation of the fidelity-realism tradeoffs induced by these parameters, and an investigation of automatic approaches for parameter selection.

\subsection{Implementation Details}
We implemented our method over two open-source diffusion models: Guided Diffusion \cite{dhariwal2021diffusion} and Stable Diffusion \cite{rombach2021highresolution}. For Guided Diffusion, we use the unconditional FFHQ \cite{karras2019style} models from \cite{choi2021ilvr} (see \supref{ss:faces}). For Stable Diffusion we use the \textit{sd-v1-4} checkpoint, except for the SVR experiments where we use the \textit{sd-v1-5} inpainting checkpoint. For all Stable Diffusion experiments we set the number of diffusion steps as $T=50$ and use DDIM sampling~\cite{song2020denoising}, for Guided Diffusion we set $T=250$ and use DDPM sampling. Inference times are affected by $R$, and are displayed in \cref{tab:inf_times}. For quantitative metrics evaluation, we use CLIP \textit{ViT-L/14@336px}.

The SVR model $\text{D}^2\text{IMNet}$ is constituted of two sub-networks, SDFNet and CamNet. Both sub-networks are trained using Adam with $5\mathrm{e}{-5}$ learning rate and $16$ batch size for $400$ epochs. This configuration remains the same for all trained models. For additional implementation details for SVR see \supref{ss:SVR model training}.

\begin{table}[b!]
\caption{Inference times of a single-image batch for varying $R$ values, evaluated on an Nvidia GeForce RTX 2080 Ti.}
{\small
\begin{tabular}{@{}cccccc@{}}
\toprule
\textbf{R} & 0 & 0.2 & 0.4 & 0.6 & 0.8 \\ \midrule
\textbf{time} [sec] & 7 & 19 & 30 & 42 & 54\\ \bottomrule
\end{tabular}
}
\label{tab:inf_times}
\end{table}

%% file: 5-limitations.tex
\section{Limitations}
\label{sec:limitations}

\begin{figure}[bt!]
\setlength{\belowcaptionskip}{-5pt}
\centerline{\includegraphics[width=0.96\linewidth]{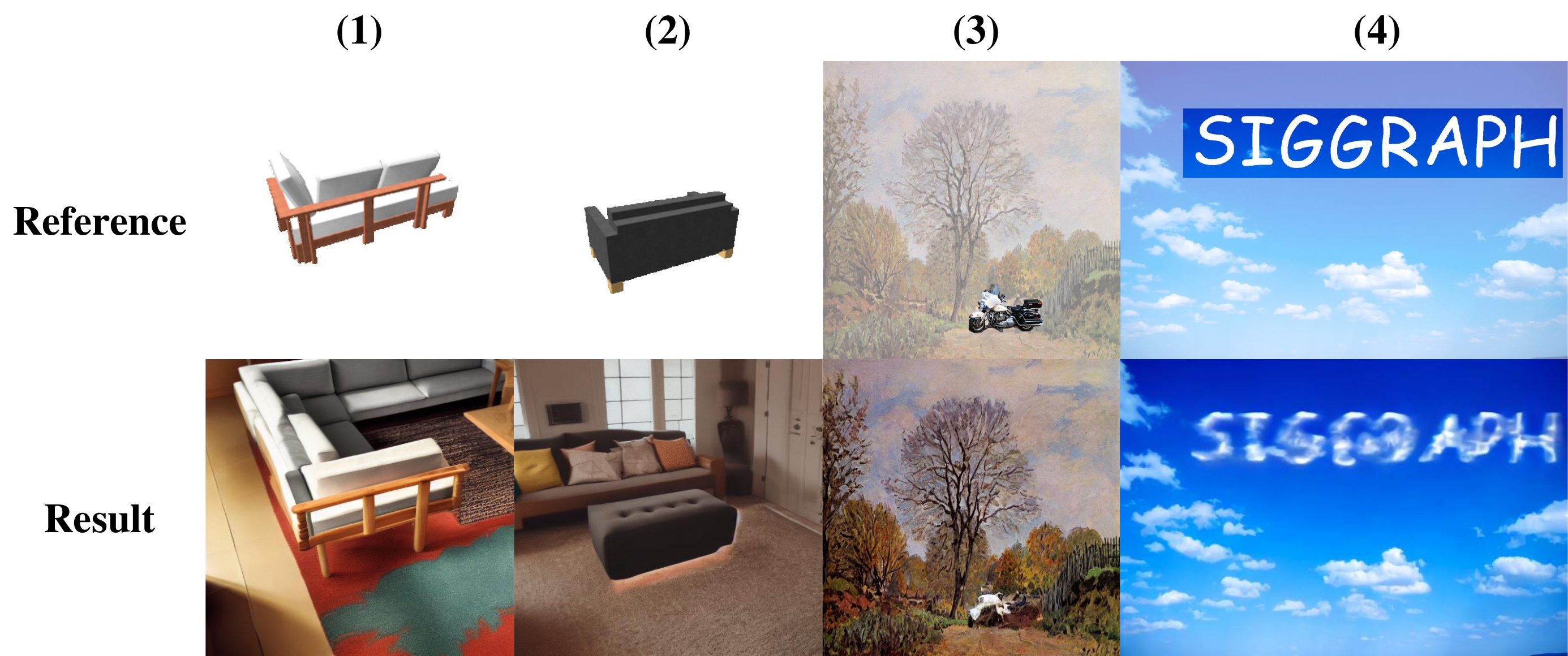}}
  \caption{\textbf{Limitations.} Failure cases of our method include: (1) Difficult and detailed objects, (2) Difficult viewpoints, making it challenging for the model to understand the semantics, (3) Small and detailed objects, (4) Out of domain content, \eg, text.}
  \label{fig:limitations}
\end{figure}

\Cref{fig:limitations} demonstrates several failure cases of our method. A primary limitation lies in applying our method to small objects. This mainly originates in our use of LDMs, which work in a low-resolution latent space and may fail to preserve fine details. We address this limitation in \supref{ss:small_objects} and propose to tackle this limitation by applying our method to cropped portions of the background, where the small inserted object represents a larger portion of the image.

Another challenge arises when the diffusion model misattributes the infused object to its semantic class. This issue often arises with challenging reference images, such as non-standard models, unnatural CG renders, or difficult views for SVR augmentation. This is notably prevalent with small, hard-to-identify images (\eg, ShapeNet airplanes). Possible mitigation strategies could involve modifying the attention layers, using methods like Paint-By-Word~\cite{balaji2022ediffi} or Prompt-to-Prompt~\cite{hertz2022prompt} for improved text and object-mask alignment. For additional discussion see \supref{ss:additional}.

Our method does not tackle all aspects of image compositing. Although it succeeds in blending the targets, and often generates proper lighting and shading that matches the scene (see \cref{fig:exp-scribbles}), it fails in generating appropriate shadows without user guidance.

Finally, as discussed throughout the paper, our method contains an inherent trade-off between fidelity and realism. This trade-off is controlled mostly by $N_{in}, T_{in}, R$ (\cref{ss:ablation}), but it may require a user to generate a grid of results for multiple configurations and manually choose the most pleasing ones. In \supref{ss:scoring} we propose a method to expedite this process by suggesting a score function based on the metrics presented in \cref{ss:metrics}.

%% file: 6-conclusions.tex
\section{Conclusions}
\label{sec:conclusions}

We presented a technique for cross-domain compositing, using pre-trained diffusion models. Our approach works at inference time, requiring no fine-tuning or optimization. Through various quantitative and qualitative studies, we demonstrated that our technique is effective for seamlessly merging partial images from distinct visual domains across different applications, including: image modification, object integration, and data augmentation for SVR.

The use of diffusion models for compositing should not be taken for granted.
Diffusion models are typically trained on the whole image, and hence they excel in global applications which consider the entire image, rather than local regions. However, models that were trained on a large number of domains have the innate capacity to compose coherent images from separate partial results. Our Cross-domain Compositing technique is based on these characteristics of large conditional diffusion models.

An interesting future work is extending cross-domain composition and its application to video. While many principles that were used for images can be easily extended to video, the maintenance of the temporal coherence along the video remains challenging.

%% file: appendix.tex
\appendix
{\Huge\textbf{Appendix}}

\section{\textbf{Mask-smoothing Implementation}} \label{ss:smoothing}
\Cref{ss:masked_ILVR} presents our mitigation for the aliasing artifacts by smoothing the blend-mask using an operator $b(M)$, in this section we expand on our implementation of this operator.

Naively blurring the mask using Gaussian blur results in an undesirable overriding of the reference in the masked regions, since the blur introduces erosion of pixels that were previously $1$ in the mask for the following equations:
\begin{align}\label{eq:sup-blend}
\phi(x;M_b)=M_b\phi_{in}(x) + (1 - M_b)\phi_{out}(x), \\
x_{t-1} = x'_{t-1}+M_t(\phi(y_{t-1})-\phi(x'_{t-1})) .
\end{align}
Instead, we would like to only blur outside the mask. We thus define $b$ as the \textit{BlurOutwards} operator, and implement it via \cref{alg:blurout}.

\begin{algorithm}[ht!]
\begin{algorithmic}
\caption{Blur Outwards}\label{alg:blurout}
    \State \textbf{Input}: Mask $M \in \{0, 1\}^{H \times W}$, number of blending pixels $p_{blend}$, smoothing function $s:[0,1] \rightarrow [0,1]$.
    \State \textbf{Output}: Blurred mask $M' \in [0, 1]^{H \times W}$.
    \State Initialize $M'=0$
   \For{$p=0,...,p_{blend}-1$}
      \State $M' \leftarrow M' + \left(s\left(\frac{p+1}{p_{blend}}\right)-s\left(\frac{p}{p_{blend}}\right)\right) \cdot M$
      \State $M \leftarrow Dilate(M)$
   \EndFor
   \State \textbf{return} $M'$
\end{algorithmic}
\end{algorithm}

\Cref{alg:blurout} blurs the mask by iteratively dilating it by one pixel, and adding up the results weighted by the smoothing function. The smoothing function $s:[0,1] \rightarrow [0,1]$ must be monotonically increasing and satisfy $s(0)=0,s(1)=1$. In practice, we define a linear smoothing function $s(x)=x$, but other functions may also be used, \eg, $softmax$. The parameter $p_{blend}>0$ controls the amount of smoothing applied to the mask, in units of pixels.

\section{\textbf{Aliasing Artifacts}} \label{ss:aliasing}
\begin{figure}[th!]
   \centerline{\includegraphics[width=1.0\linewidth]{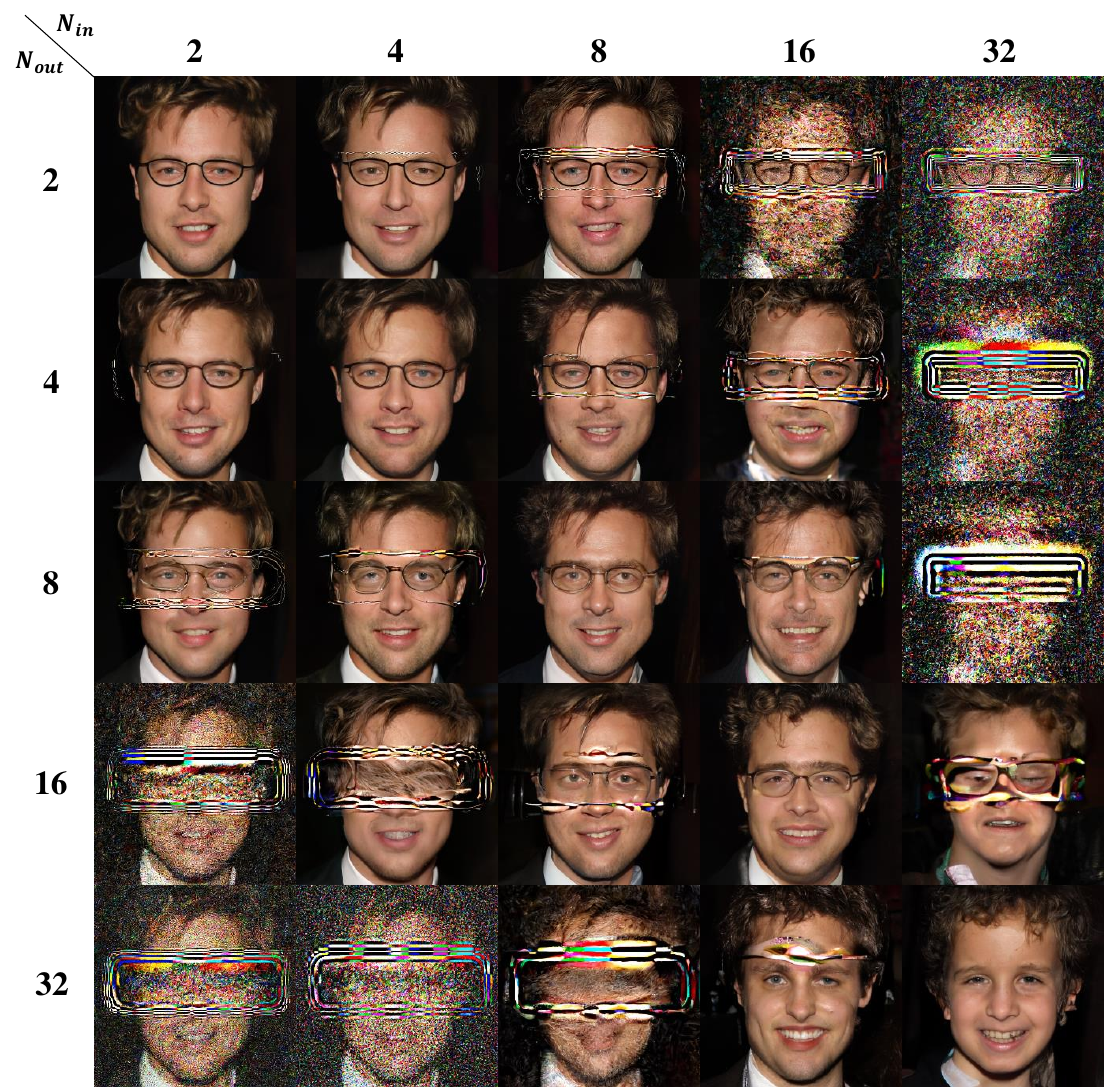}}
  \caption{\textbf{Artifacts as a variation of $N_{in}, N_{out}$.} As the difference between the scaling factors increases, creating sharper edges between regions, the artifact effect increases.}
  \label{fig:sup-artifactsN}
\end{figure}

\begin{figure}[tbh!]
   \centerline{\includegraphics[width=1.0\linewidth]{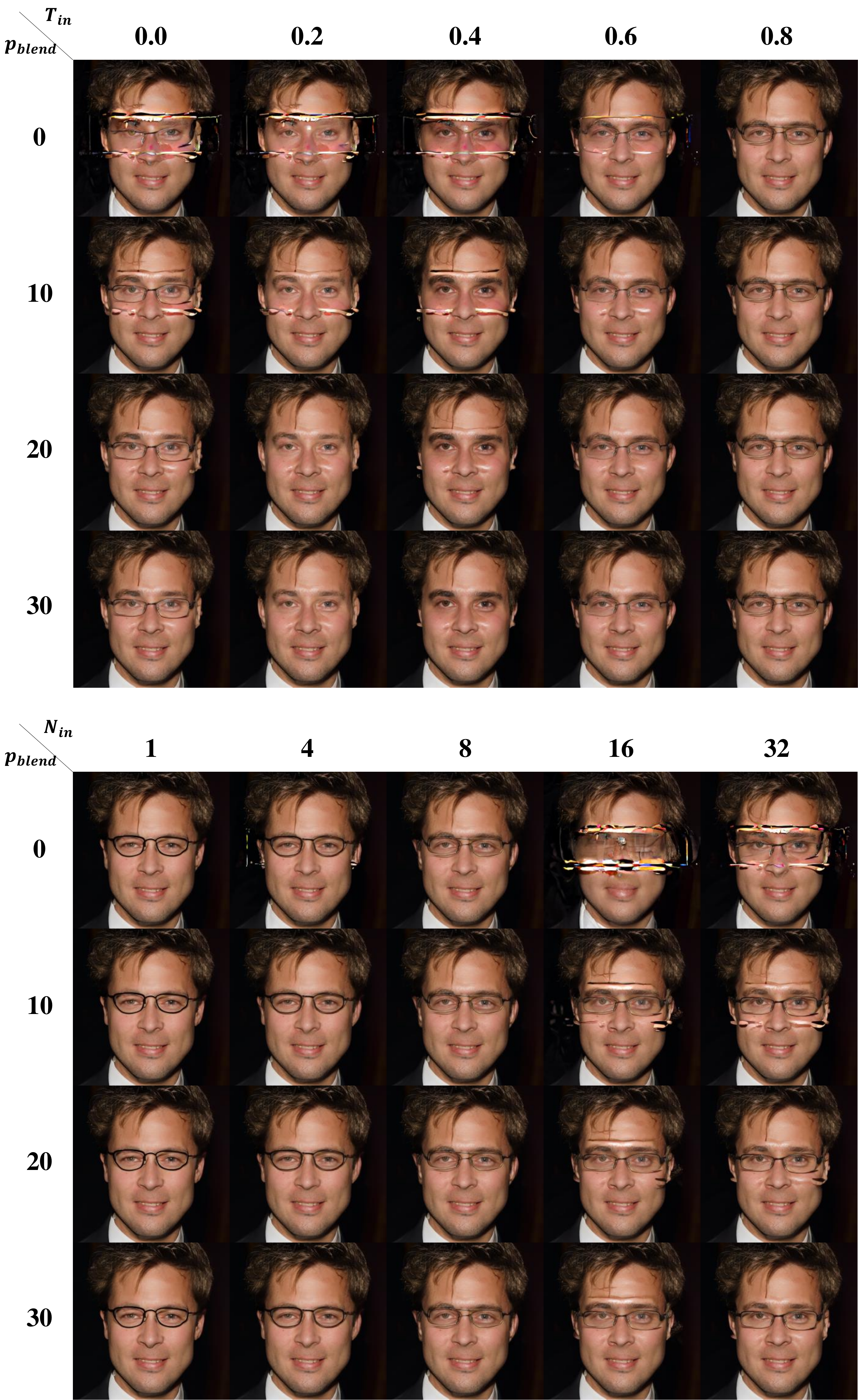}}
  \caption{\textbf{Blending mitigation.} Top: Effect of varying $T_{in}$ (top) and $N_{in}$ (bottom) on the artifacts, mitigated by increasing $p_{blend}$. Other parameters are fixed at $T_{out}=0.8, N_{out}=8$, and $N_{in}=8$ (top), $T_{in}=0.8$ (bottom).}
  \label{fig:sup-artifacts_blend}
\end{figure}

This section provides a deeper dive into the effects of the aliasing artifacts mentioned in \cref{ss:masked_ILVR}. \Cref{fig:sup-artifactsN} demonstrates the effect of varying $N_{in}, N_{out}$ on the strength of the observed aliasing artifacts. As the difference between the scaling factors increases, in effect increasing the strength of the low-pass filtering, sharper edges are introduced to the intermediate images. This causes an aliasing effect which results in wave-like artifacts around the edges. A similar effect happens upon setting a large difference between $T_{in}$ and $T_{out}$, which inflates the sharp edges for a longer period during the intermediate steps. Our blending mitigation is presented in \cref{fig:sup-artifacts_blend}, establishing a clear relationship between the $N_{in}, N_{out}, T_{in}, T_{out}$ and the blending strength required to diminish the effect.

\section{\textbf{Face Modification}} \label{ss:faces}
\begin{figure}[h!]
   \centerline{\includegraphics[width=1.0\linewidth]{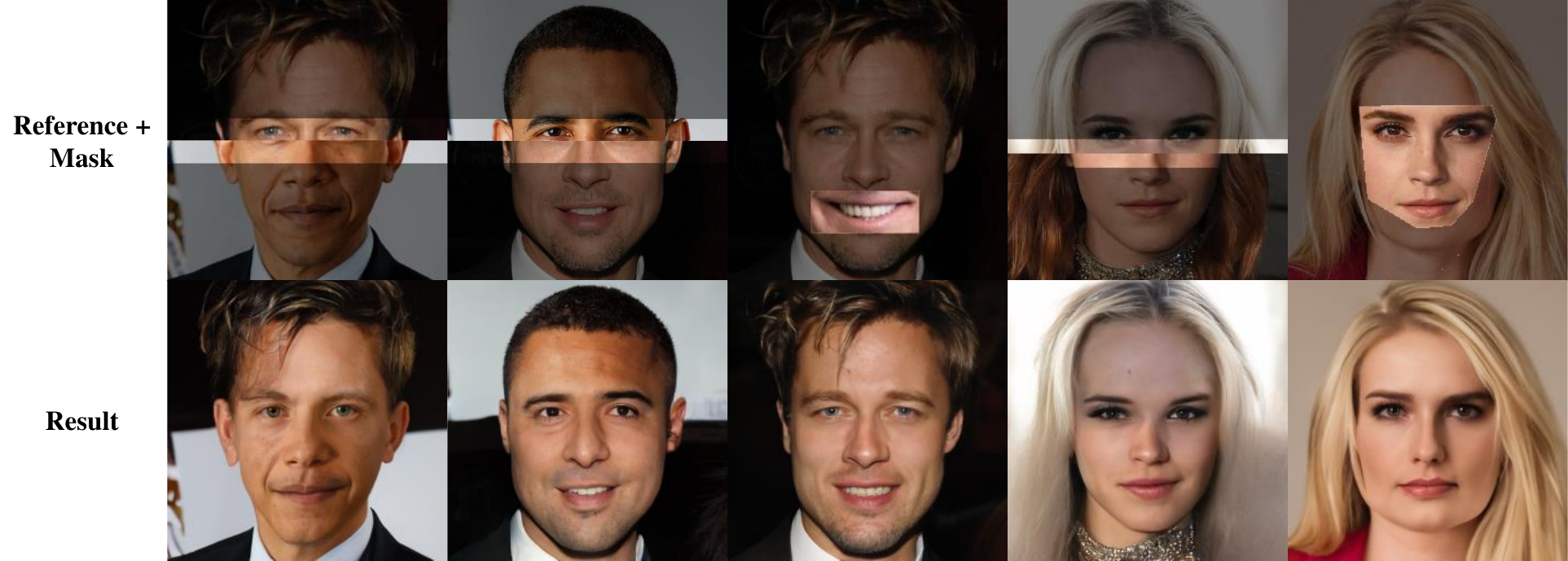}}
  \caption{\textbf{Face modification examples.}}
  \label{fig:sup-faces}
\end{figure}

In addition to the various applications displayed in the main paper, our method may also be used for editing facial features. This is done by employing diffusion models trained on portrait datasets, \eg, FFHQ. \Cref{fig:sup-faces} displays results on this task, proving the ability to blend faces, add smiles, or even perform face-swapping. These results also demonstrate how the strength of our method relies on the expressiveness capabilities of the diffusion model. A model trained on a uni-domain dataset (as in this case) has a better ability to perform challenging edits, due to its capability to keep intermediate results in-domain.

\begin{figure}[t!]
   \centerline{\includegraphics[width=1.0\linewidth]{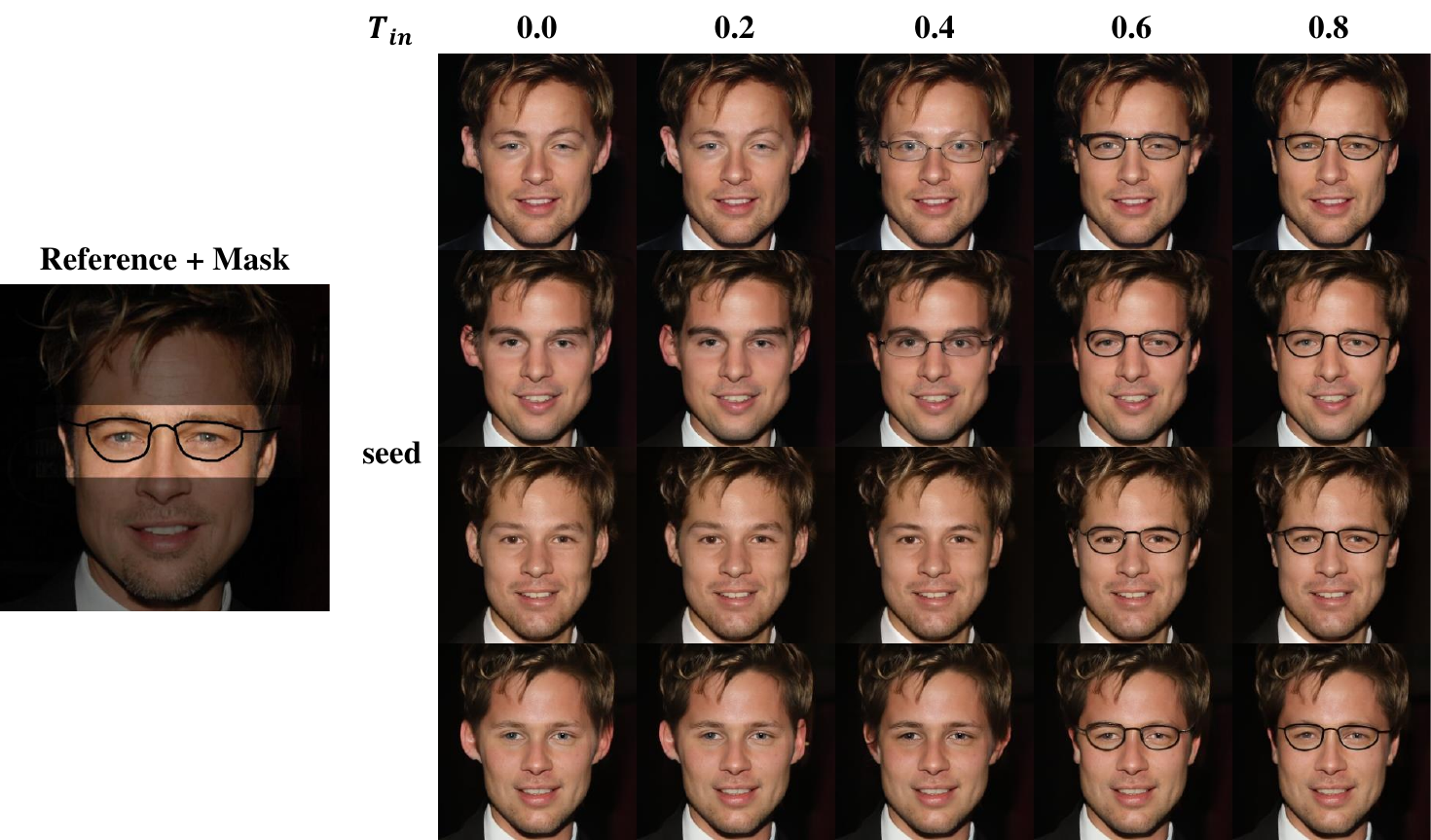}}
  \caption{\textbf{Modifying both regions simultaneously.} In this experiment, parameters were fixed at $N_{in}=N_{out}=2,T_{out}=0.8,p_{blend}=30$, while $T_{in}$ and the seed vary. Our method allows changing different image regions with different levels of control, presenting diversity of the results.}
  \label{fig:sup-faces_blend}
\end{figure}

In most applications displayed in this paper, control parameters of the outer region were fixed at $N_{out}=1,T_{out}=1$ as to locally modify only the masked region, and keep the rest of the image unchanged. In \cref{fig:sup-faces_blend}, results are shown for a different case, setting $N_{out}=2,T_{out}=0.8$ and thus allowing slight changes to the outer region as well, while simultaneously editing the inner region to a different level. These results also demonstrate the diversity introduced by increasing $N$, as presented in ILVR.

\section{\textbf{Analysis of Fidelity-Realism Trade-off}} \label{ss:metrics-analysis}
As discussed in \cref{sec:method}, our method contains an inherent trade-off between \textit{fidelity} to the reference image and \textit{realism} to the target domain. We present a deeper dive into this trade-off by analyzing metrics from \cref{ss:metrics}. We focus on the scribble-based editing task, using 24 example images. We generate 300 results for each image, comprised of 10 $T_{in}$ values, 6 $N_{in}$ values, and 5 $R$ values. We then calculate LPIPS as the fidelity score and CLIP Directional Similarity score as the realism score.

\begin{figure}[tbh!]
\centerline{\includegraphics[width=0.95\linewidth]{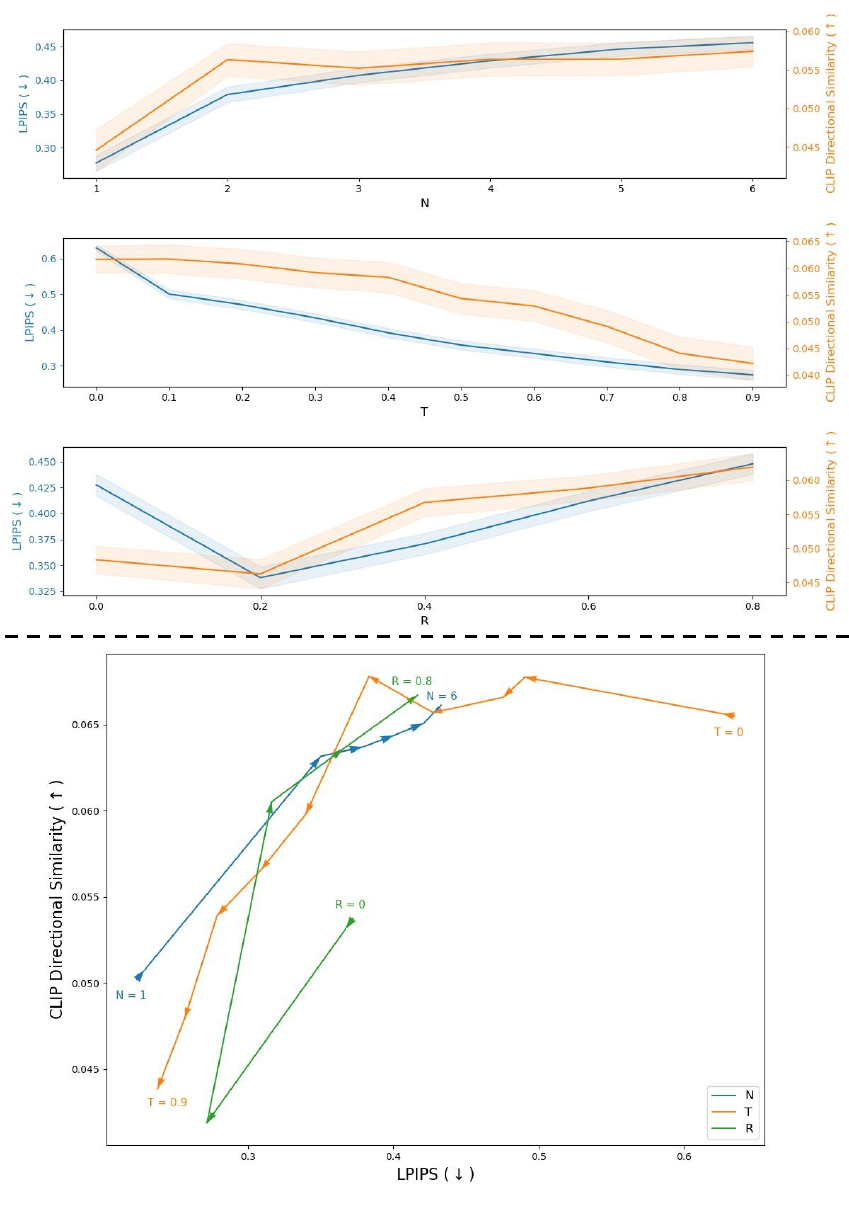}}
  \caption{\textbf{Deeper analysis of fidelity-realism trade-off using.} Top: Each graph shows the effect of different parameters on fidelity (LPIPS) and realism (CLIP Directional Similarity) scores. For each graph we fix a certain parameter, and average the other two amongst all images. Bottom: Each curve demonstrates how the average scores change together with the change of the parameter.}
  \label{fig:tradeoff}
\end{figure}

\Cref{fig:tradeoff} presents an analysis of these scores. First, it is noticeable that the fidelity and realism scores are clearly anti-correlated, as one decays when the other improves. Second, as explained in \cref{ss:ablation}, the method's parameters have control over this trade-off. A higher $N_{in}$ passes less information from the reference image to the prediction, and thus results in lower fidelity, but allows for better realism. It is worth noting that the realism score reaches some plato for $N_{in}$, which coincides with our observation that high $N_{in}$ values have little diversity from one another. Adversely, increasing $T_{in}$ performs the frequency-overriding process for more steps, resulting in higher fidelity but lower realism. Interestingly, both scores are not monotonic for $R$ in one point ($R = 0.2$), but overall increasing $R$ improves the realism of the image by applying more diffusion steps, at the cost of fidelity.

\section{\textbf{Semi-Automatic Scoring Method}} \label{ss:scoring}
A limitation of our method is the lack of a single set of parameters that result in satisfying results for all inputs, impelling the user to generate a grid of results, and then manually choose the most appealing result out of all images. To mitigate this issue, we propose a scoring method based on the metrics presented in \cref{ss:metrics}, and the analysis discussed in \cref{ss:metrics-analysis}.

\begin{figure}[tb!]
   \centerline{\includegraphics[width=0.97\linewidth]{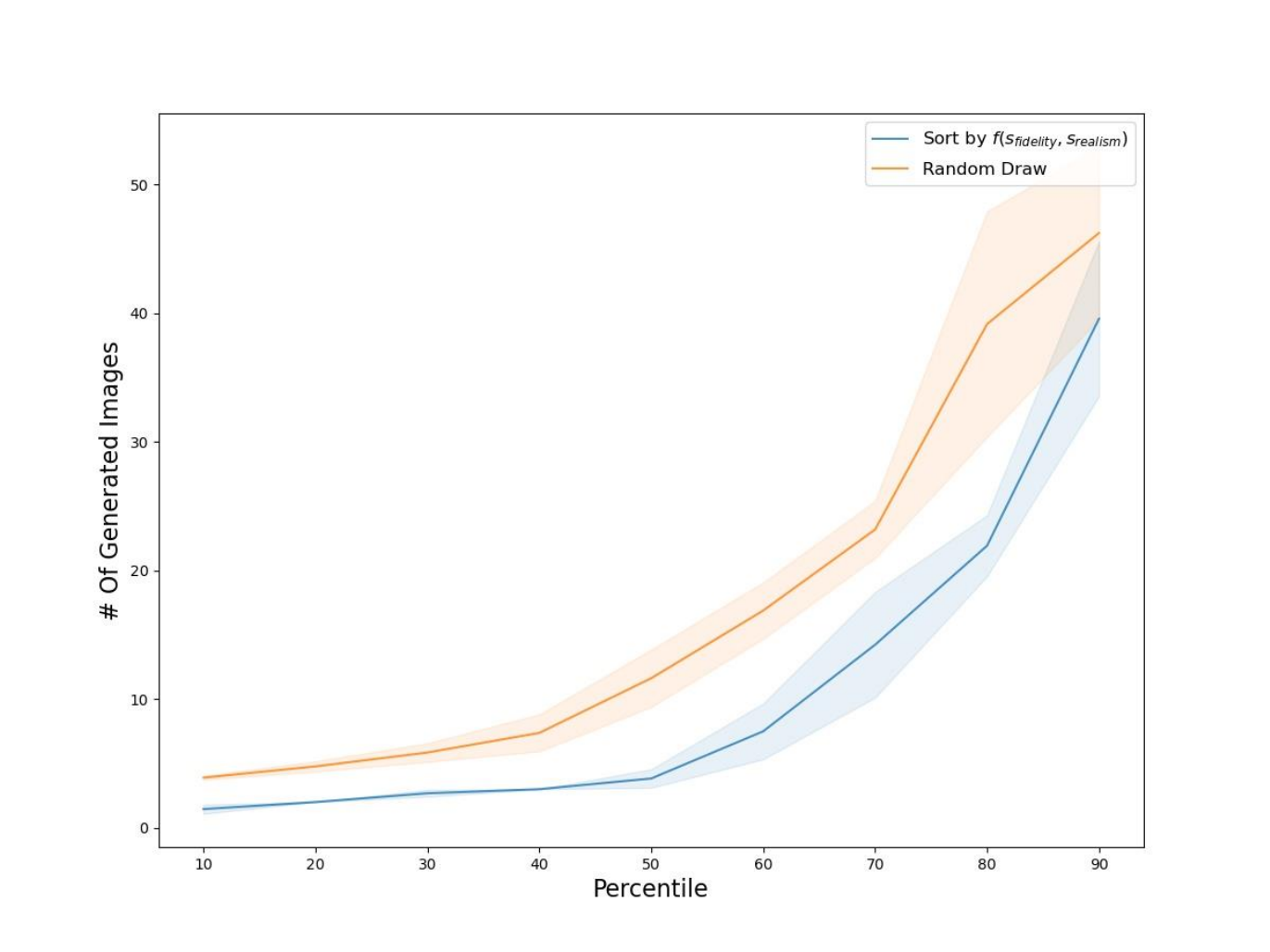}}
  \caption{\textbf{Comparison of percentile curves for scoring method vs. random sampling method.} Percentiles of the minimal number of samples required to find a pleasing result using both random configurations and our sorting method. Error bars are calculated by bootstrapping. Our score function provides an improved sorting method to find pleasing results.}
  \label{fig:scoring}
\end{figure}

Let $s_{fidelity}, s_{realism}$ denote the fidelity and realism scores, and defined here as LPIPS and CLIP Directional Similarity respectively. We first normalize both scores to $[0, 1]$ (and take the negative of LPIPS since lower is better). The score function is defined as
\begin{align}\label{eq:scoring}
f(s_{fidelity}, s_{realism}) = \frac{1}{1 + \lambda}\left(s_{fidelity}+\lambda s_{realsim}\right) .
\end{align}
We set $\lambda=2$ with the justification that realism is somewhat preferable over fidelity. To analyze this scoring method we focus on the scribble-based editing task. For each input image, we manually choose a set of the most appealing results out of the 300 generated. We sort all 300 results using the score function and report the first appearance of one of the selected appealing results by applying this sorting method. In \cref{fig:scoring} we compare the percentile curves of this sorting method and a random sorting method. For a random sort, the mean first appearance of an appealing image would be $\frac{300}{k}$, where $k$ is the number of selected appealing images for a certain reference image. Using our proposed score function to sort results may help the user to find a pleasing result quicker, by examining results in the order of decreasing scores.

\section{\textbf{Small Objects}} \label{ss:small_objects}
\begin{figure}[t!]
\centerline{\includegraphics[width=1.0\linewidth]{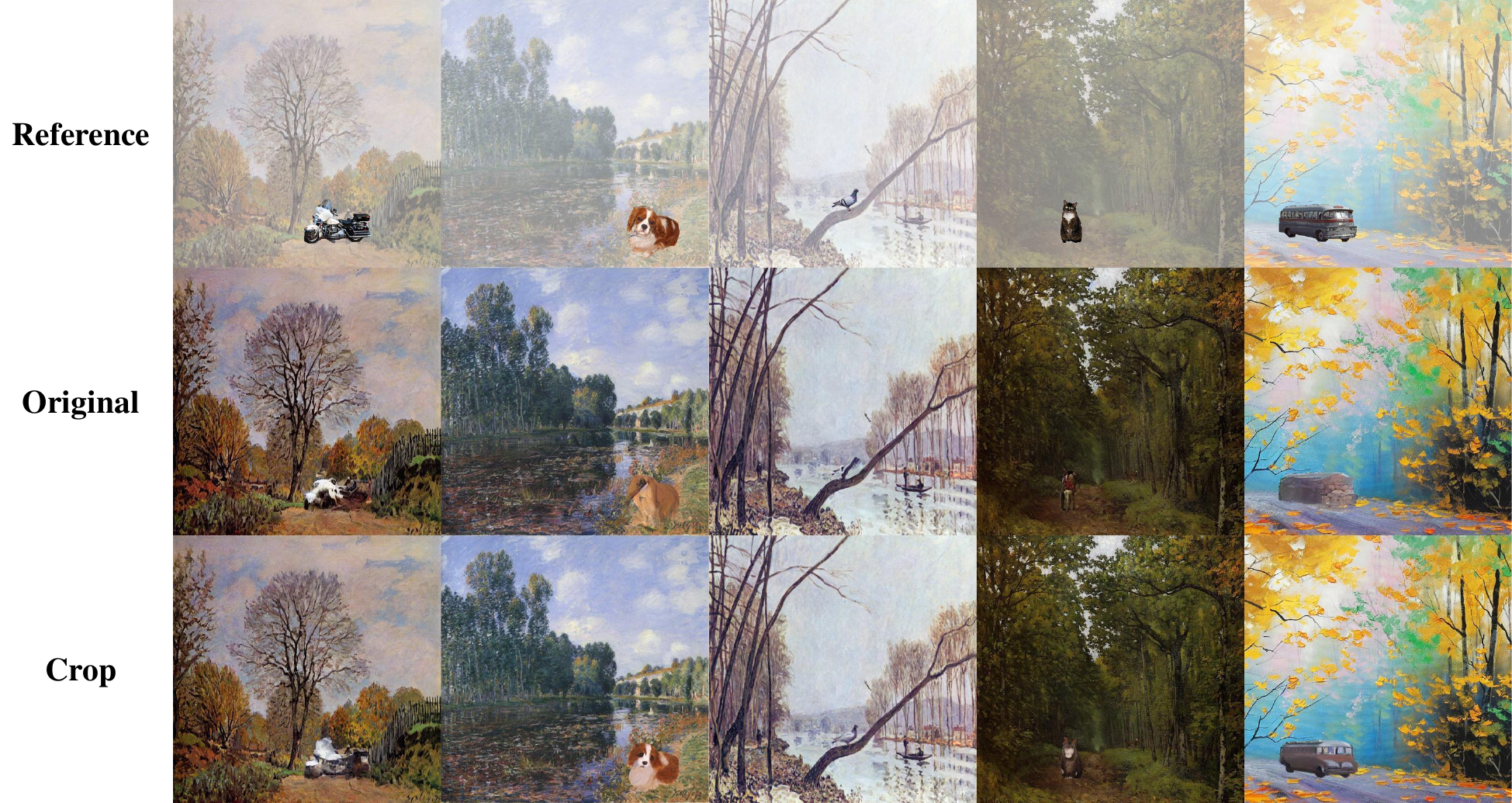}}
  \caption{\textbf{Small objects mitigation using crops.} Top: reference image. Middle: original result using our method, small objects tend to become smeared. Bottom: using cropped images as reference results in fine-detail preservation, leading to improved results.}
  \label{fig:crop}
\end{figure}

To relieve our method's difficulty with processing small objects, we propose a simple method in the preprocessing phase. Instead of using the entire original image as a reference, we crop a bounding box two times the size of a tight bounding box around the object, upsample and resize it, process it via our method (\cref{sec:method}), resize the result back to the original size, and finally paste it back in the original image. This process assists in preserving enough information from the small reference object in the low-resolution latent space, as opposed to using the original-sized object which results in an insufficient latent representation of the object in LDMs. Results using this method are presented in \cref{fig:crop}, demonstrating improved results with better preservation of the reference image's finer details and semantics.

\begin{figure}[t!]
\centerline{\includegraphics[width=1.0\linewidth]{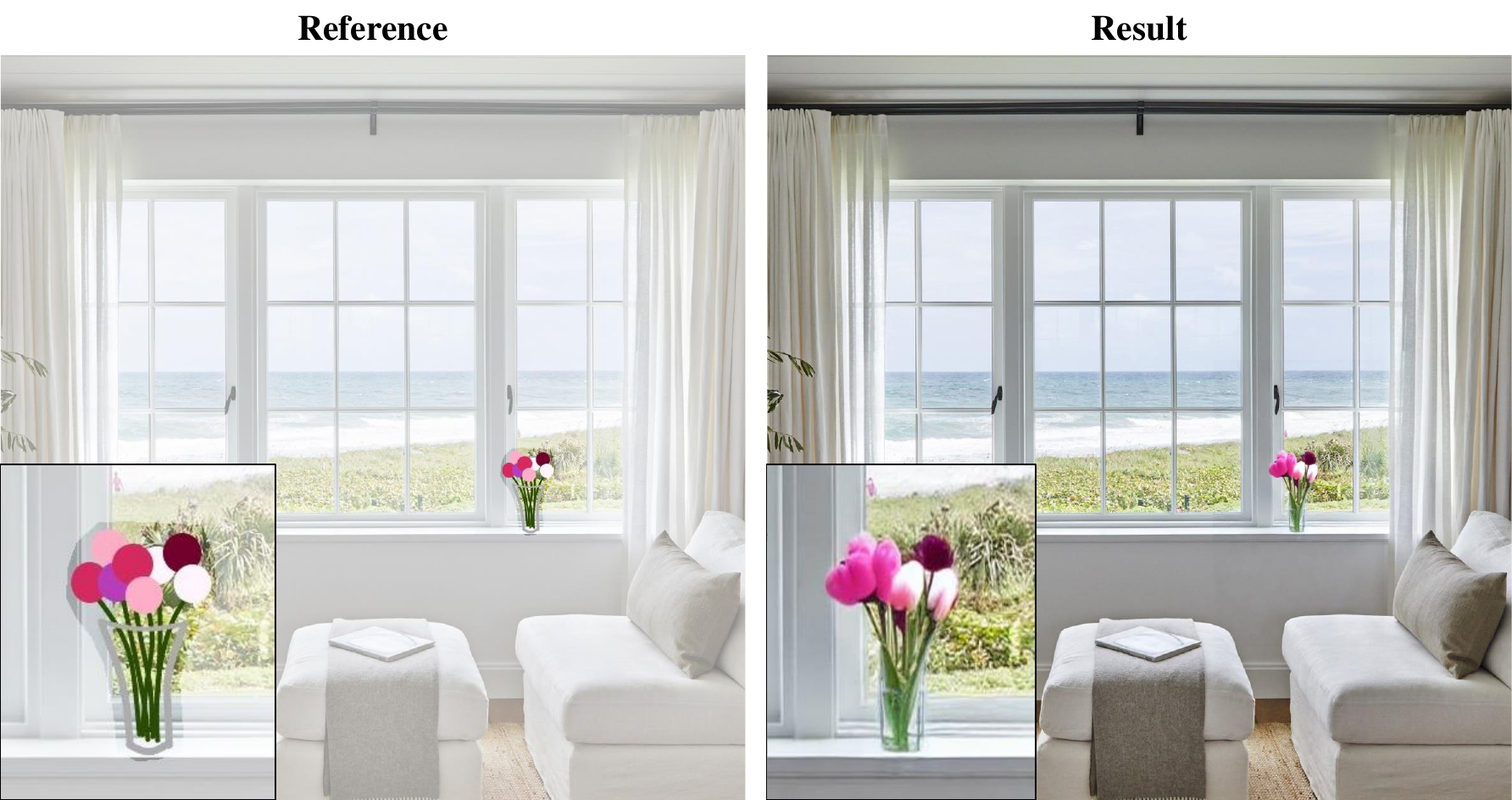}}
  \caption{\textbf{Small objects in large images.} An example of an edit on a $1688 \times 1688$ image. Prompt: \textit{a photo of a flower vase on a window}, configuration: $N_{in}=2$, $T_{in}=0.8$, $R=0.8$.}
  \label{fig:crop-large}
\end{figure}

This preprocessing method has the bonus effect of unlocking the ability to work on arbitrarily sized images, even those that are too large for the diffusion model's capacity. To do so, one may crop a small bounding box around the desired editing region, process it, and seamlessly paste it back into the original image. An example of this on a $1688\times1688$ image is shown in  \cref{fig:crop-large}. Note that this is applicable only as long as the editing region is not larger than the model's capacity.

\section{\textbf{Timestep Schedule}} \label{ss:repaint}
\begin{figure}[h!]
   \centerline{\includegraphics[width=1.0\linewidth]{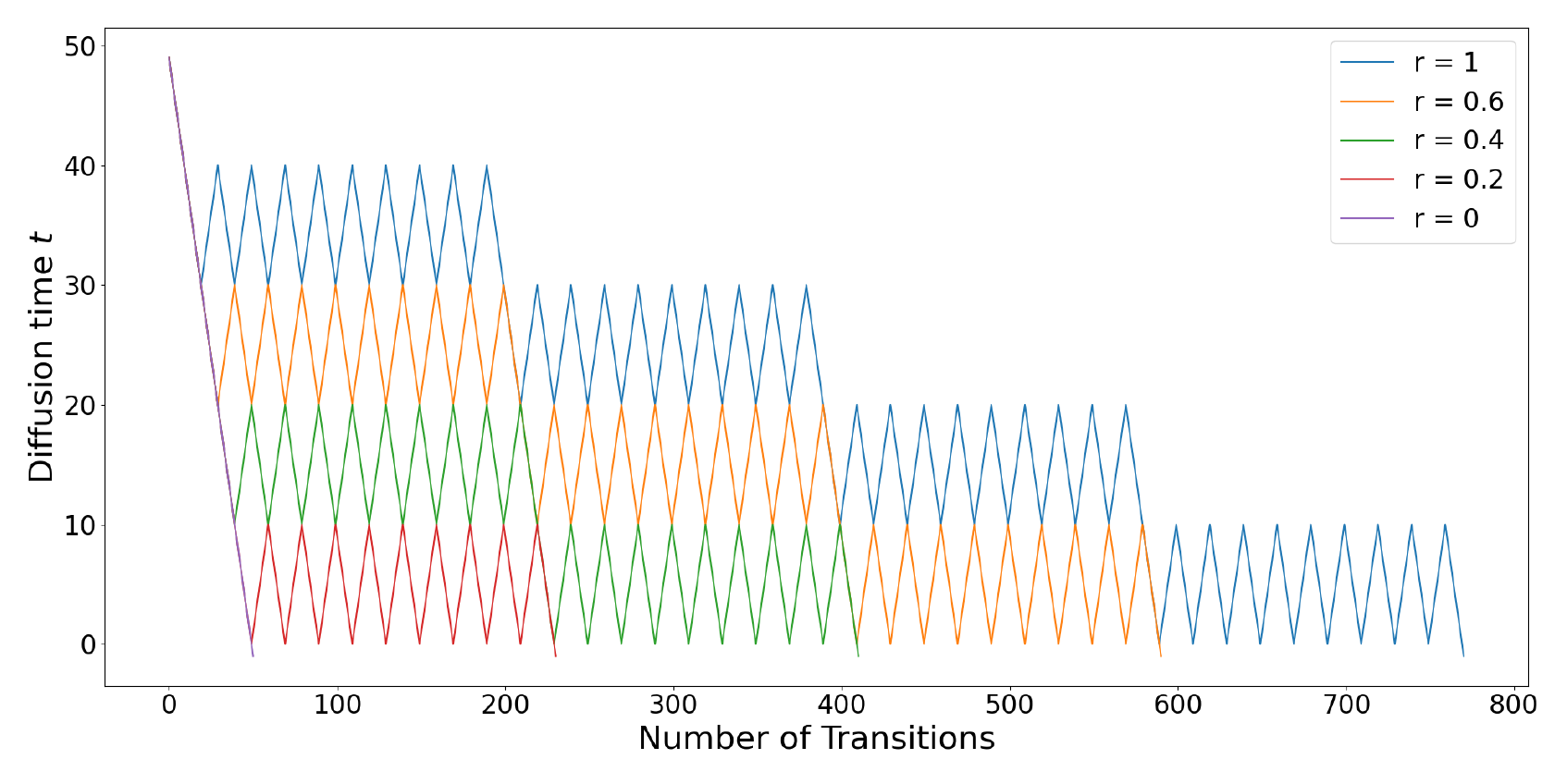}}
  \caption{\textbf{RePaint Schedule.} Changing the value of $R$ sets the timestep from which resampling is applied. $R=0$ reduces to a normal backward-diffusion schedule.}
  \label{fig:sup-repaint}
\end{figure}

As described in \cref{ss:masked_ILVR}, we add control over transferring semantic and style details into the edited region by controlling the scheduling of the backward-diffusion. RePaint offers to perform resampling during the prediction (renoise and perform another forward-pass of the diffusion model), effectively increasing the receptive field of the model. Our control is achieved by setting the timestep in which resampling starts, defined as $R$ in the paper. \Cref{fig:sup-repaint} displays several schedules achieved by altering $R$. $R=0$ reduces to no resampling at all, falling back to a normal backward-diffusion schedule. More resampling gives rise to deeper harmonization between the object and the background but may cause corruption of the object.

\section{\textbf{SVR Data Augmentation Details}} \label{ss:SVR model training}
We utilized a dataset of rendered images from \cite{DISN} for augmenting ShapeNet object data with background features, intended for training our SVR models. This dataset encompasses 36 rendered views across 20 categories of ShapeNet models. Three separate sets of SVR experiments were conducted, each with unique data augmentation procedures involving distinct prompts and subsampled rendered views. The specific methodologies for each experiment are detailed in \cref{tab: sup augmentation details}.

To construct a copy-and-paste dataset, where background augmentation is achieved through naïve image composition, we employed pretrained diffusion models to generate background images which were then composited with ShapeNet objects. The generation of background images was guided by customized text prompts tailored specifically for each object category. More than 3,500 background images were produced for each category.

\begin{table*}[t!]
\caption{Implementation details of background augmentation on rendered views of ShapeNet objects.}
\begin{tabular}{@{}ccccc@{}}
\toprule
Category               & \# Objects            & \# Views & $T_{in}$           & Prompt                                                                                                                                                             \\ \midrule
\multirow{2}{*}{Sofa}  & \multirow{2}{*}{3173} & 6        & 0.25,0.75,1.0 & \multirow{2}{*}{"A photograph of a sofa in a living room with windows and a rug"}                                                                                  \\
                       &                       & 36       & 0.5           &                                                                                                                                                                    \\ \midrule
\multirow{2}{*}{Chair} & \multirow{2}{*}{6778} & 8        & 0.25,0.75,1.0 & \multirow{2}{*}{\begin{tabular}[c]{@{}c@{}}"Photograph of a chair sitting on the floor in a room, interior design, \\ natural light, photorealistic"\end{tabular}} \\
                       &                       & 20       & 0.5           &                                                                                                                                                                    \\ \midrule
Table                  & 8436                  & 7        & 0.5           & \begin{tabular}[c]{@{}c@{}}"Photograph of a table in the room, natural light, interior design, \\ realistic indoor scenes, cozy room"\end{tabular}                 \\ \bottomrule
\end{tabular}
\label{tab: sup augmentation details}
\end{table*}

\section{\textbf{2D IOU Implementation}} \label{ss:2DIoU}
\begin{figure}[t!]
   \centerline{\includegraphics[width=1.0\linewidth]{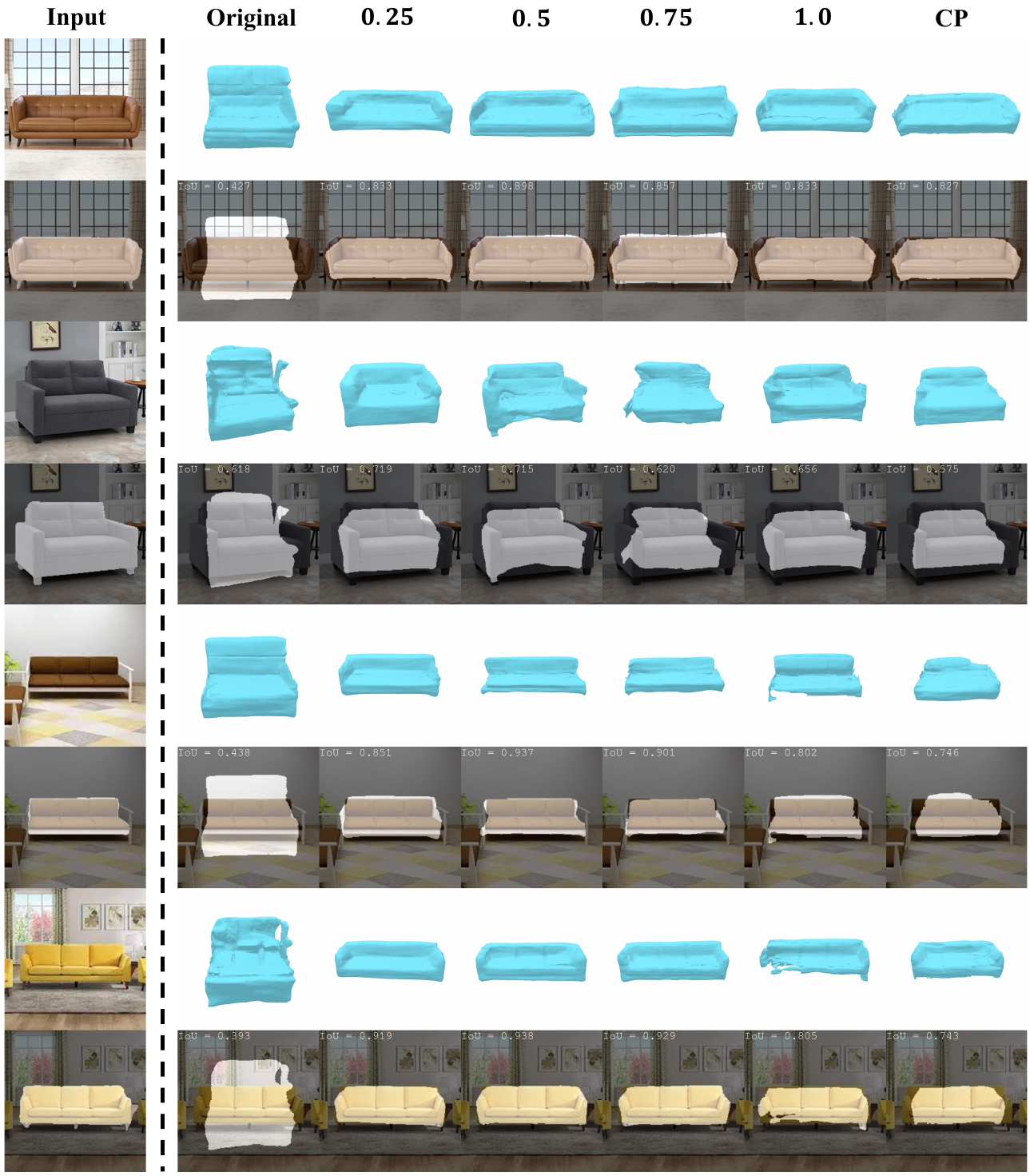}}
  \caption{\textbf{2D IoU evaluation samples.} Odd rows: Input images and the 3D reconstruction results by models trained on respective dataset. Even rows: Mask overlays of the rendered object silhouette on the input image. The measured 2D IoU between object silhouette and input image mask is annotated at the top left corner of the image.}
  \label{fig:2D IoU}
\end{figure}

In lack of a sufficient 3D ground-truth for evaluating results on internet-sourced images, we propose to apply 2D IoU as a proxy which allows us to take maximal leverage of the available information. This metric is determined over a 2D mask on internet-sourced input images and reconstructed model silhouette. We extract object masks from input images through off-the-shelf segmentation model SEEM \cite{zou2023segment}, and render SVR results under camera pose predicted by CamNet. The input image mask and model silhouette are aligned with their geometric center, then the intersection over union between these two masks is determined. A sample 2D IoU measured over several internet-sourced images is shown in \cref{fig:2D IoU}. This metric evaluates how faithful the reconstruction results are with respect to the input view-point. 

\section{\textbf{Additional Results}} \label{ss:additional}
\begin{figure*}
   \centerline{\includegraphics[width=1.0\linewidth]{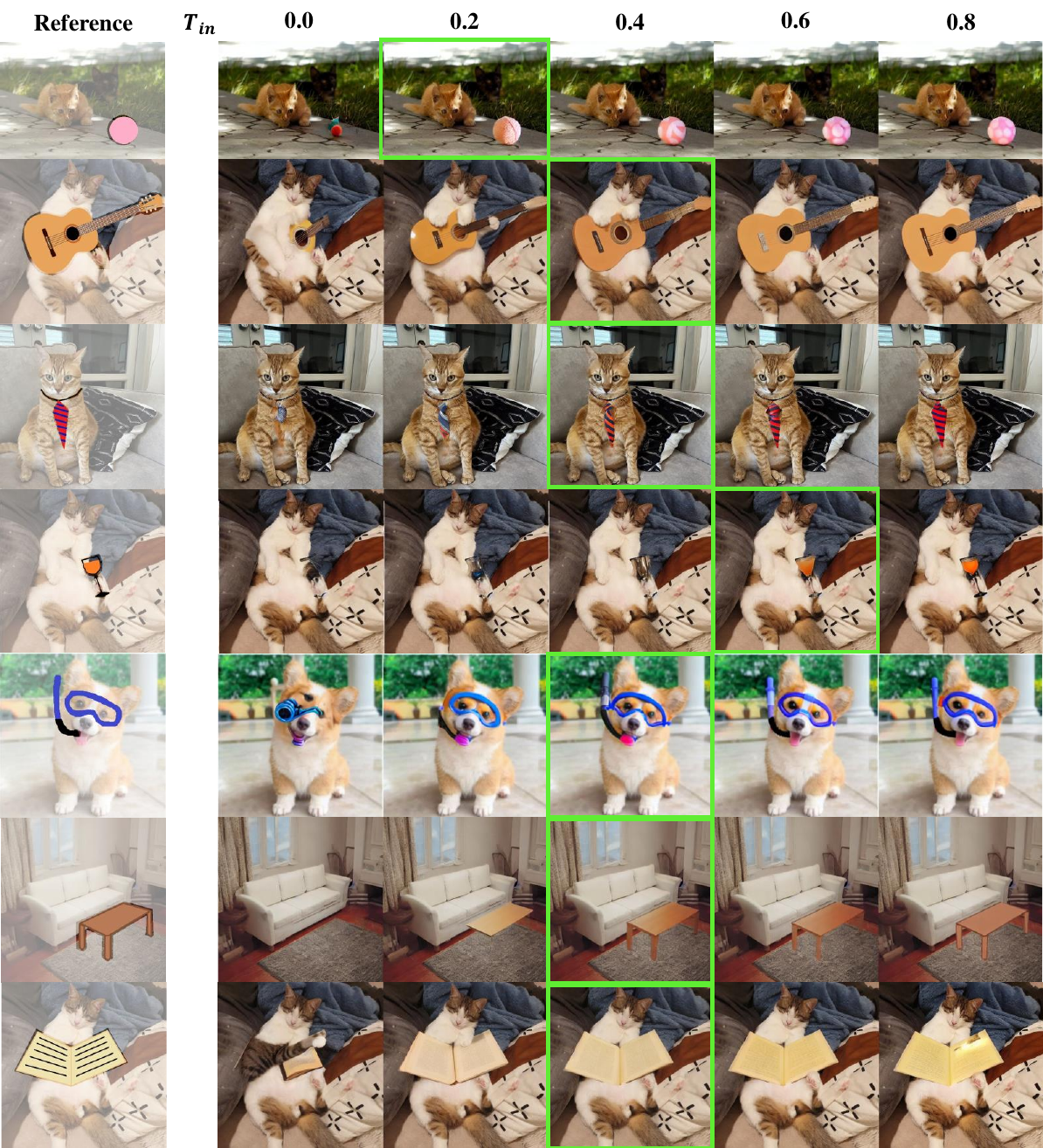}}
  \caption{Additional scribble images generated using our method, for various $T_{in}$ values, the most appealing results (as rated by users) are highlighted. Text prompts from top to bottom are: \textit{A cat playing with a yarn ball}, \textit{A playing guitar}, \textit{A cat reading a book}, \textit{A cat wearing a tie}, \textit{A cat drinking a cocktail}, \textit{A corgi with a snorkel mask}, \textit{A living room with a sofa and a table}, \textit{A cat reading a book}.}
  \label{fig:sup-scribbles}
\end{figure*}

\begin{figure*}
   \centerline{\includegraphics[width=1.0\linewidth]{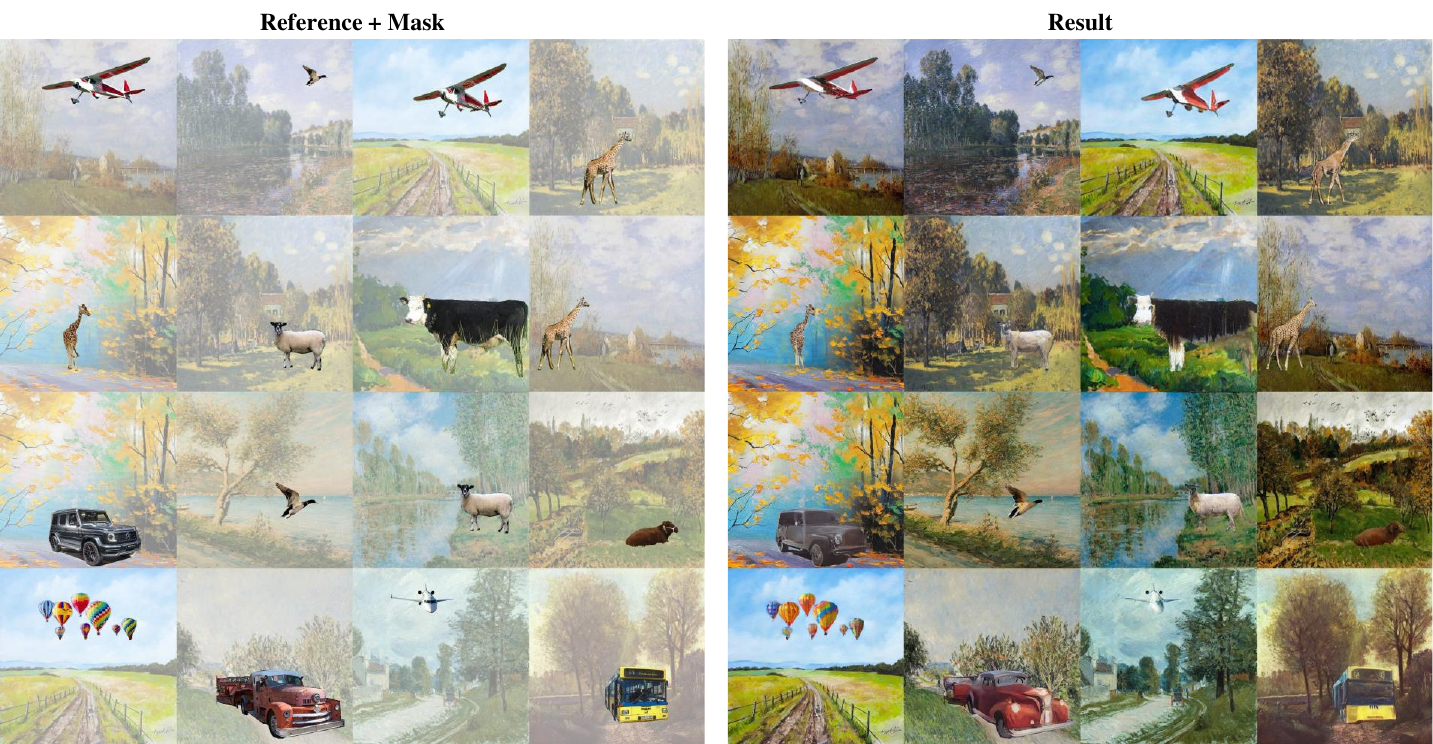}}
  \caption{Additional object immersion results edited using our method, all generated using $T_{in}=0.5, N_{in}=2, T_{out}=1, N_{out}=1, R=0.2$.}
  \label{fig:sup-immersion}
\end{figure*}

\begin{figure*}
   \centerline{\includegraphics[height=0.8\textheight]{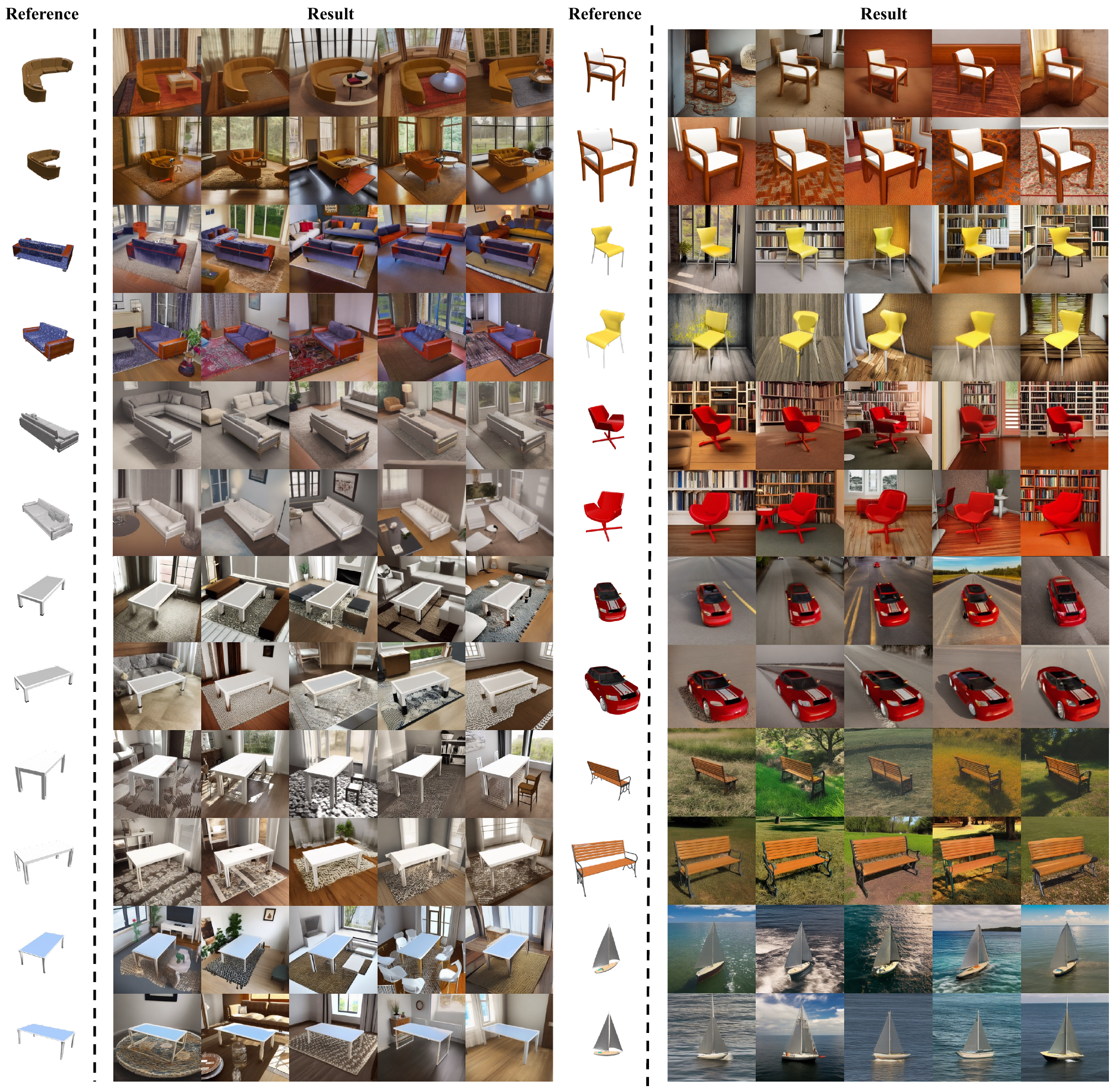}}
  \caption{Additional augmentation results of various sofa, table, chair, car, bench and watercraft images and views, all generated using $T_{in}=0.5, N_{in}=1, T_{out}=0, R=0$ and a pretrained inpainting model. Text prompt for sofa: \textit{Photograph of a sofa in the living room with window and a rug}; chair: \textit{Photograph of a chair sitting on the floor in a room, interior design, natural light, photorealistic} and \textit{Photograph of a chair in the study near a bookshelf}; table: \textit{Photograph of a table in the room, natural light, interior design, realistic indoor scenes, cozy room}; car: \textit{Photograph of a car driving on the road}; bench: \textit{Photograph of a bench in the park with grasses in a sunny day}; watercraft: \textit{Photograph of a watercraft in the ocean under blue sky and white clouds}.}
  \label{fig:sup-all-cat}
\end{figure*}

In this section, we supply additional quantitative and qualitative results for the various applications displayed in the paper. In \cref{fig:sup-scribbles,fig:sup-immersion,fig:sup-all-cat} we provide additional results generated by our method, demonstrating the robustness and generalization capabilities of our local editing method. \Cref{fig:sup-scribbles} shows additional scribble + text prompt examples, establishing how setting an intermediate $T_{in}$ aids in transforming the scribble to become more realistic, and that a \naive attempt of inpainting with no guidance ($T_{in}=0$) is often insufficient.

\Cref{fig:sup-immersion} presents examples for Object Immersion, including a few failure cases, mostly originating from small objects. In \cref{fig:sup-all-cat} additional SVR augmentation examples on various object categories are displayed, demonstrating our method's ability to generate realistic and diverse images merely from a single image of the object. An interesting failure case is observed in the second row of the left grid, where the diffusion model sometimes chooses to break the single L-sofa into two regular sofas, probably due to its prior knowledge from training.

\begin{figure}[!t]
   \centerline{\includegraphics[height=0.4\textheight]{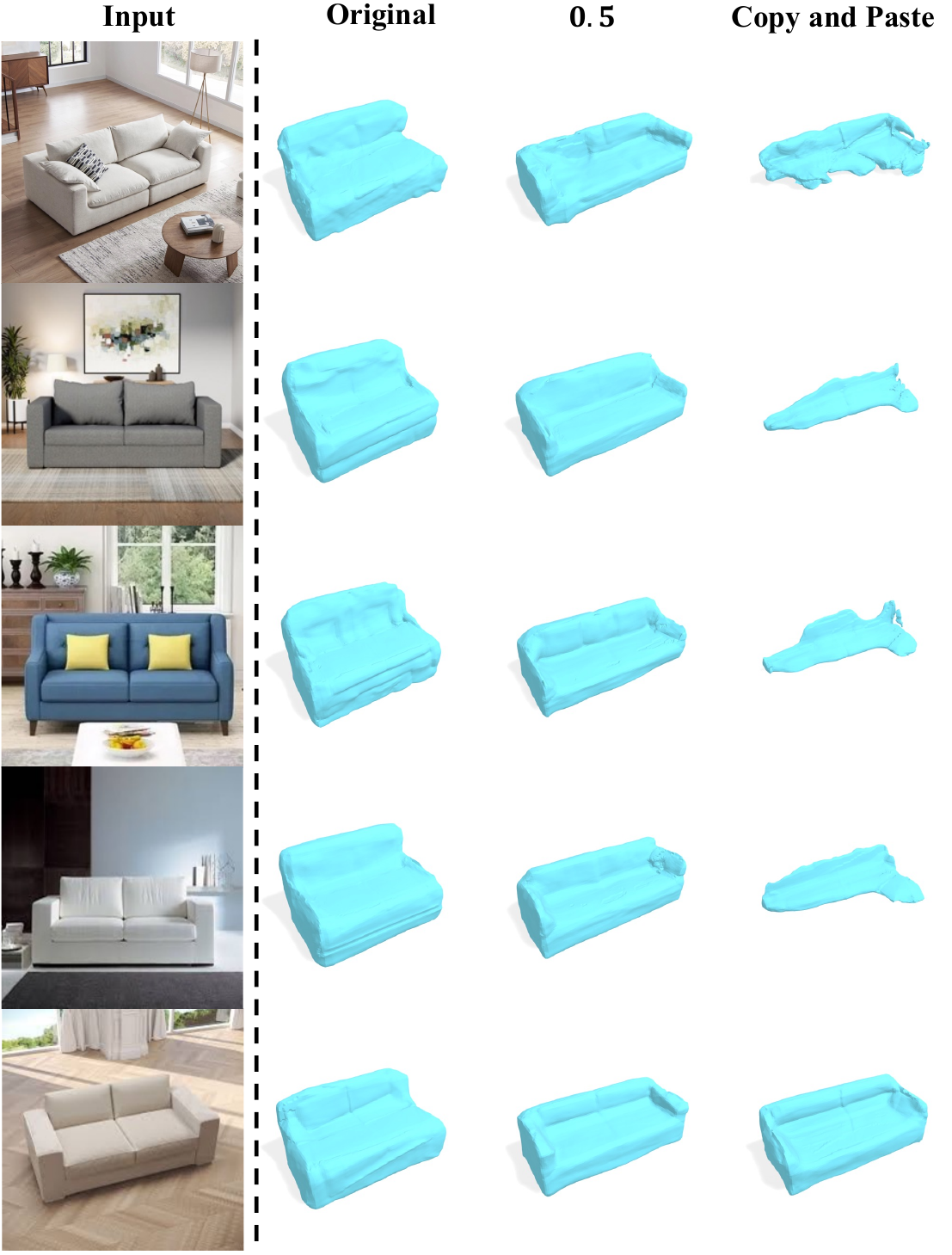}}
  \caption{Additional SVR results tested on internet-sourced sofa images. The SVR model is trained on an augmented sofa dataset with $T_{in} = 0.5$, compared with original and copy-paste baselines.}
  \label{fig:svr-sofa}
\end{figure}

\begin{figure}[!t]
   \centerline{\includegraphics[height=0.4\textheight]{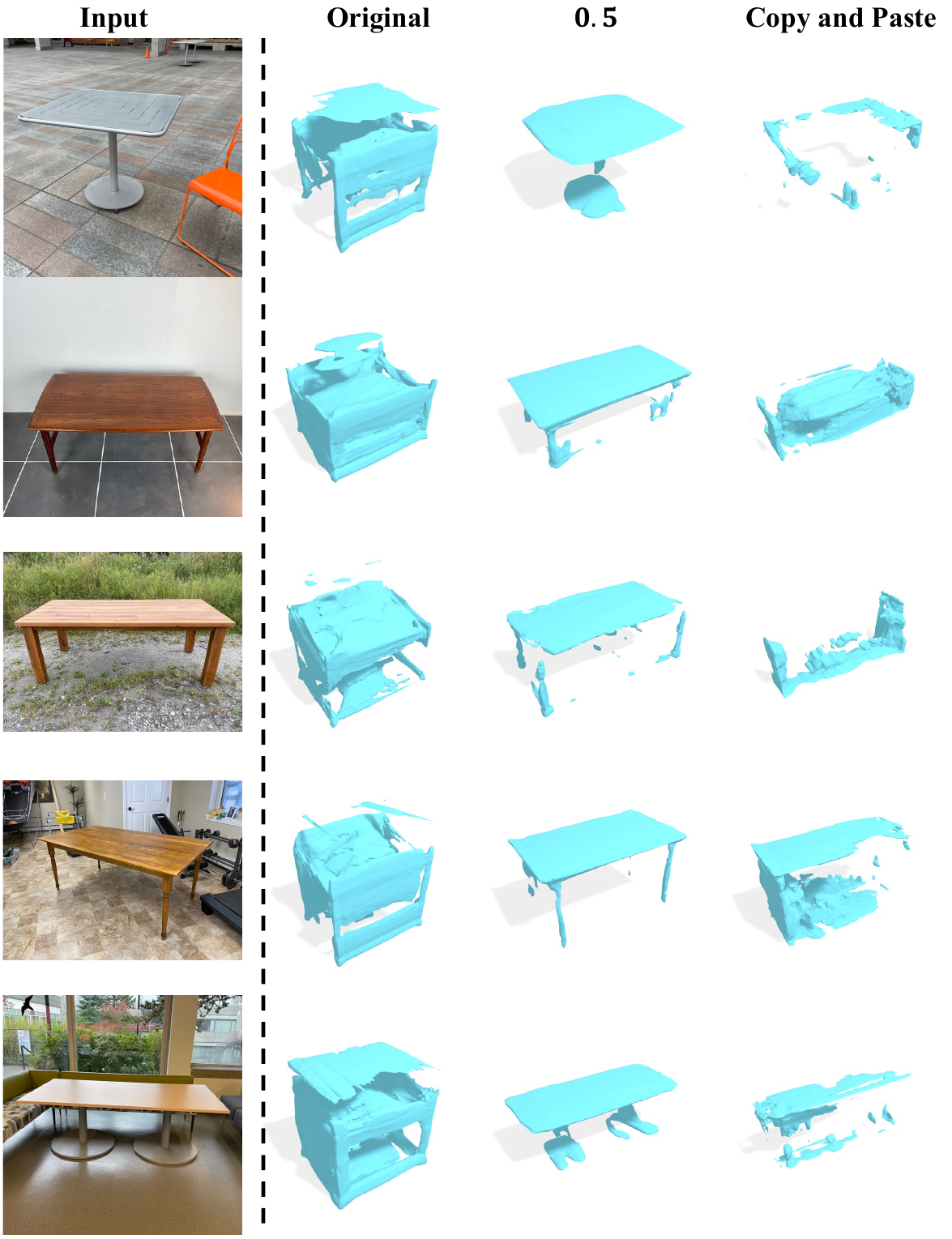}}
  \caption{Additional SVR results tested on in-the-wild table images. The SVR model is trained on augmented table dataset with $T_{in} = 0.5$, compared with original and copy-paste baselines.}
  \label{fig:svr-table}
\end{figure}

\begin{figure}[!t]
   \centerline{\includegraphics[height=0.4\textheight]{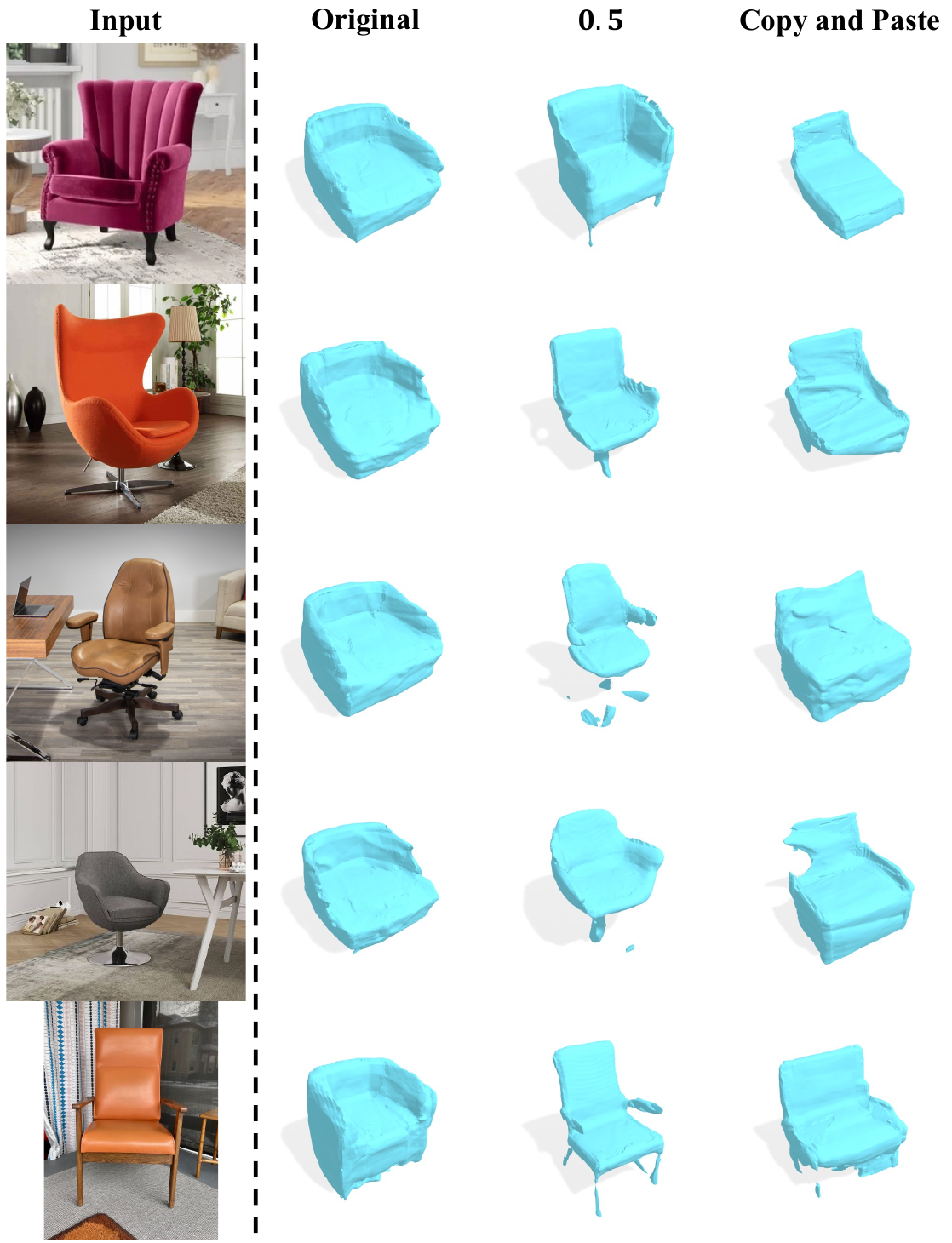}}
  \caption{Additional SVR results tested on internet-sourced chair images. The SVR model is trained on an augmented chair dataset with $T_{in} = 0.5$, compared with original and copy-paste baselines.}
  \label{fig:svr-chairs}
\end{figure}

Additional SVR models are trained on additional object categories with extended views. The quantitative measurements are demonstrated in \cref{tab: sup SVR eval}, further establishing the superiority of our augmentation method. The number of views is extended to 36 for sofas, 20 for chairs, and 7 for tables. The augmentation pipeline follows from \cref{ss:SVR model training}. The models are tested on 100 internet-sourced images for each category, and additional qualitative results are shown in \cref{fig:svr-sofa,fig:svr-chairs,fig:svr-table}. Results using our augmentation method appear to have better coherency to the input object and better overall structure.

While our method outperforms the baselines in all cases, the SVR models often struggle with the reconstruction of thin elements in input images. This is due to the inherent challenge of implicitly recognizing and extracting these fine details from in-the-wild images, and the fact that our augmentation pipeline occasionally overlooks these parts due to their small size. 

Another notable limitation lies in the range of input image viewpoints. The views rendered from ShapeNet provide limited variations in camera poses, a stark contrast to the virtually infinite camera pose possibilities in-the-wild images can present. While our augmentation of background aids in bridging the real-to-sim domain gap, developing a model with robust camera pose tolerance extends beyond the scope of this work.

However, these limitations can be partially mitigated by preprocessing in-the-wild images, such as by positioning the object centrally and/or applying random scaling. We hope to stimulate further research in these areas to improve upon these limitations and advance the capabilities of single-view reconstruction models.

\begin{table}[t!]
\caption{2D IoU evaluated on models trained on different datasets with extended views.}
\begin{tabular}{@{}cccc@{}}
\toprule
      & Original & Ours ($T_{in} = 0.5$)   & CP    \\ \midrule
Sofa (36 views)  & 0.514    & \textbf{0.601} & 0.524 \\
Chair (20 views) & 0.416    & \textbf{0.581} & 0.491 \\
Table (7 views) & 0.280    & \textbf{0.417}  & 0.334 \\ \bottomrule
\end{tabular}
\label{tab: sup SVR eval}
\end{table}

\clearpage

